\documentclass[nonatbib]{article} 
\usepackage{iclr2024_conference,times}

\usepackage{amsmath,amsfonts,bm}

\def\eqref#1{equation~\ref{#1}}

\def\1{\bm{1}}

\DeclareMathAlphabet{\mathsfit}{\encodingdefault}{\sfdefault}{m}{sl}
\SetMathAlphabet{\mathsfit}{bold}{\encodingdefault}{\sfdefault}{bx}{n}

\DeclareMathOperator*{\argmin}{arg\,min}

\usepackage{url}
\usepackage{amsfonts}
\usepackage{textcomp}
\usepackage{amssymb}
\usepackage[linesnumbered,ruled,vlined]{algorithm2e}
\usepackage{relsize}

\usepackage{float}
\usepackage{multirow}
\usepackage{caption}

\usepackage[utf8]{inputenc} 
\usepackage[T1]{fontenc}    
\usepackage{hyperref}       
\usepackage{url}            
\usepackage{booktabs}       
\usepackage{amsfonts}       
\usepackage{nicefrac}       
\usepackage{microtype}      
\usepackage{xcolor}         

\usepackage{amsmath}
\usepackage{amssymb}
\usepackage{amsthm}
\usepackage{array}
\usepackage{epsfig}
\usepackage{graphicx}
\usepackage{xy}

\usepackage{times}
\usepackage{epsfig}
\usepackage{graphicx}
\usepackage{amsmath}
\usepackage{amssymb}
\usepackage{multirow}
\usepackage[ruled,vlined]{algorithm2e}
\usepackage{subcaption}
\usepackage{comment}
\usepackage[export]{adjustbox}

\newtheorem{theorem}{Theorem}
\newtheorem{definition}{Definition}

\newtheorem*{theorem1}{Theorem~\ref{theorem1}}

\newcommand{\norm}[1]{\left\lVert#1\right\rVert}

\title{Mode-Aware Continual Learning for Conditional Generative Adversarial Networks}

\author{Cat P. Le, Juncheng Dong, Ahmed Aloui, Vahid Tarokh\\
Department of Electrical and Computer Engineering, Duke University\\
}

\iclrfinalcopy 

\begin{document}

\maketitle

\begin{abstract}
The main challenge in continual learning for generative models is to effectively learn new target modes with limited samples while preserving previously learned ones. To this end, we introduce a new continual learning approach for conditional generative adversarial networks by leveraging a mode-affinity score specifically designed for generative modeling. First, the generator produces samples of existing modes for subsequent replay. The discriminator is then used to compute the mode similarity measure, which identifies a set of closest existing modes to the target. Subsequently, a label for the target mode is generated and given as a weighted average of the labels within this set. We extend the continual learning model by training it on the target data with the newly-generated label, while performing memory replay to mitigate the risk of catastrophic forgetting. Experimental results on benchmark datasets demonstrate the gains of our continual learning approach over the state-of-the-art methods, even when using fewer training samples.
\end{abstract}


\section{Introduction}
\label{intro}
Artificial intelligence (AI) for generative tasks has made significant progress in recent years, and we have seen remarkable applications, such as ChatGPT~\citep{openai}, DALL-E~\citep{vaswani2021dalle}, and deepfake~\citep{westerlund2019emergence}. However, most of these methods~\cite {wang2018transferring, varshney2021cam, zhai2019lifelong, le2020supervised, seff2017continual} lack the ability to learn continuously, which remains a challenging problem in developing AI models that can match human's continuous learning capabilities. This challenge is particularly difficult when the target data is limited or scarce~\citep{varshney2021cam, zhai2019lifelong, seff2017continual}. In this scenario, the goal is to learn to generate new images using an extensive model that is trained on all previous tasks. Most continual learning methods focus on preventing the models from forgetting the existing tasks, but many learning restrictions are often enforced on learning new tasks, leading to poor performance. To efficiently learn the new task, relevant knowledge can be identified and utilized. To this end, various knowledge transfer approaches have been introduced, resulting in significant breakthroughs in many applications, including natural language processing~\citep{vaswani2017attention, devlin2018bert, howard2018universal, le2023improving, brown2020language}, and image classification~\citep{elaraby2022conditional, guo2019spottune, ge2017borrowing, le2022task, cui2018large, azizi2021big}. These techniques enable models to leverage past experiences, such as trained models, and hyper-parameters to improve their performance on the new task, emulating how humans learn and adapt to new challenges (e.g., learning to ride a motorcycle is less challenging for someone who already knows how to ride a bicycle). It is also essential to identify the most relevant task for knowledge transfer when dealing with multiple learned tasks. Irrelevant knowledge can be harmful when learning new tasks~\citep{le2022task, standley2020tasks}, resulting in flawed conclusions (e.g., misclassifying dolphins as fish instead of mammals could lead to misconceptions about their reproduction).


In this paper, we propose a \textit{Discriminator-based Mode Affinity Score} (dMAS) to evaluate the similarity between generative tasks and present a new few-shot continual learning approach for the conditional generative adversarial network (cGAN)~\citep{mirza2014conditional}. Our approach allows for seamless and efficient integration of new tasks' knowledge by identifying and utilizing suitable information from previously learned modes. Here, each mode corresponds to a generative task. 
Our framework first evaluates the similarity between the existing modes and the new task using dMAS. It enables the identification of the most relevant modes whose knowledge can be leveraged for quick learning of the target task while preserving the knowledge of the existing modes. To this end, we add an additional mode to the generative model to represent the target task. This mode is assigned a class label derived from the labels of the relevant modes and the computed distances. Moreover, we incorporate memory replay~\citep{robins1995catastrophic, chenshen2018memory} to prevent catastrophic forgetting. 

Extensive experiments are conducted on the MNIST~\citep{lecun2010mnist}, CIFAR-10~\citep{krizhevsky2009learning}, CIFAR-100~\citep{krizhevsky2009learning}, and Oxford Flower~\citep{nilsback2008automated} datasets to validate the efficacy of our proposed approach. We empirically demonstrate the stability and robustness of dMAS, showing that it is invariant to the model initialization. Next, we apply this measure to the continual learning scenario. Here, dMAS helps significantly reduce the required data samples, and effectively utilize knowledge from the learned modes to learn new tasks. We achieve competitive results compared with baselines and the state-of-the-art approaches, including individual learning~\citep{mirza2014conditional}, sequential fine-tuning~\citep{wang2018transferring}, multi-task learning~\citep{standley2020tasks}, EWC-GAN~\citep{seff2017continual}, Lifelong-GAN~\citep{zhai2019lifelong}, and CAM-GAN~\citep{varshney2021cam}. The contributions of our paper are summarized below:

\begin{itemize}
    \item We propose a new discriminator-based mode-affinity measure (dMAS), to quantify the similarity between modes in conditional generative adversarial networks.
    \item We provide theoretical analysis (i.e., Theorem~\ref{theorem1}) and empirical evaluation to demonstrate the robustness and stability of dMAS.
    \item We present a new mode-aware continual learning framework using dMAS for cGAN that adds the target mode to the model via the weighted label from the relevant learned modes.
\end{itemize}

\section{Related Works}
\label{related-work}
Continual learning involves the problem of learning a new task while avoiding catastrophic forgetting~\citep{kirkpatrick2017overcoming, mccloskey1989catastrophic, carpenter1987massively}. It has been extensively studied in image classification~\citep{kirkpatrick2017overcoming, achilledynamic, rebuffi2017icarl, verma2021efficient, zenke2017continual, wu2018memory, singh2020calibrating, rajasegaran2020itaml}. In image generation, previous works have addressed continual learning for a small number of tasks or modes in GANs~\citep{mirza2014conditional}. These approaches, such as memory replay~\citep{wu2018memory}, have been proposed to prevent catastrophic forgetting~\citep{zhai2019lifelong, cong2020gan, rios2018closed}. However, as the number of modes increases, network expansion~\citep{yoon2017lifelong, xu2018reinforced, zhai2020piggyback, mallya2018packnet, masana2020ternary, rajasegaran2019random} becomes necessary to efficiently learn new modes while retaining previously learned ones. Nevertheless, the excessive increase in the number of parameters remains a major concern. 

The concept of task similarity has been widely investigated in transfer learning, which assumes that similar tasks share some common knowledge that can be transferred from one to another. However, existing approaches in transfer learning~\citep{kirkpatrick2017overcoming, silver2008guest,  finn2016deep, mihalkova2007mapping, niculescu2007inductive, luolabel, Razavian:2014:CFO:2679599.2679731, 5288526, zamir2018taskonomy, chen2018coupled} mostly focus on sharing the model weights from the learned tasks to the new task without explicitly identifying the closest tasks. In recent years, several works~\citep{le2022task, zamir2018taskonomy, le2021task, le2022fisher, le2021improved, aloui2022causal, pal2019zeroshot, dwivedi2019, achille2019task2vec, wang2019neural, pmlr-v119-standley20a} have investigated the relationship between image classification tasks and applied relevant knowledge to improve overall performance. However, for the image generation tasks, the common approaches to quantify the similarity between tasks or modes are using common image evaluation metrics, such as Fr\'echet Inception Distance (FID)~\citep{heusel2017gans} and Inception Score (IS)~\citep{salimans2016improved}. While these metrics can provide meaningful similarity measures between two distributions of images, they do not capture the state of the GAN model and therefore may not be suitable for transfer learning and continual learning. For example, a GAN model trained to generate images for one task may not be useful for another task because this model is not well-trained, even if the images for both tasks are visually similar. In continual learning for image generation~\citep{wang2018transferring, varshney2021cam, zhai2019lifelong, seff2017continual}, however, mode-affinity has not been explicitly considered. Although some prior works~\citep{zhai2019lifelong, seff2017continual} have explored fine-tuning cGAN models (e.g., WGAN~\citep{arjovsky2017wasserstein}, BicycleGAN~\citep{zhu2017toward}) with regularization techniques, such as Elastic Weight Consolidation (EWC)~\citep{kirkpatrick2017overcoming}, or the Knowledge Distillation~\citep{hinton2015distilling}, they did not focus on measuring mode similarity or selecting the closest modes for knowledge transfer. Other approaches use different assumptions such as global parameters for all modes and individual parameters for particular modes~\citep{varshney2021cam}. Their proposed task distances also require a well-trained target generator, making them unsuitable for real-world continual learning scenarios.


\section{Proposed Approach}
\subsection{Mode Affinity Score}
\label{method}
We consider a conditional generative adversarial network (cGAN) that is trained on a set $S$ of source generative tasks, where each task represents a distinct class of data. The cGAN consists of two key components: the generator $\mathcal{G}$ and the discriminator $\mathcal{D}$. Each source generative task $a \in S$, which is characterized by data $X_a$ and its labels $y_a$, corresponds to a specific \textit{mode} in the well-trained generator $\mathcal{G}$. Let $X_b$ denote the incoming target data. Here, we propose a new mode-affinity measure, called Discriminator-based Mode Affinity Score (dMAS), to showcase the complexity involved in transferring knowledge between different modes in cGAN. This measure involves computing the expectation of Hessian matrices computed from the discriminator's loss function. To calculate the dMAS, we begin by feeding the source data $X_a$ into the discriminator $\mathcal{D}$ to compute the corresponding loss. By taking the second-order derivative of the discriminator's loss with respect to the input, we obtain the source Hessian matrix. Similarly, we repeat this process using the target data $X_b$ as input to the discriminator, resulting in the target Hessian matrix. These matrices offer valuable insights into the significance of the model's parameters concerning the desired data distribution. The dMAS is defined as the Fr\'echet distance between these Hessian matrices.

\begin{definition}[Discriminator-based Mode Affinity Score]\label{dTAS}
Consider a well-trained cGAN with discriminator $\mathcal{D}$ and the generator $\mathcal{G}$ that has $S$ learned modes. For the source mode $a \in S$, let $X_a$ denote the real data, and $\tilde{X}_a$ be the generated data from mode $a$ of the generator $\mathcal{G}$. Given $X_b$ is the target real data, $H_{a}, H_{b}$ denote the expectation of the Hessian matrices derived from the loss function of the discriminator $\mathcal{D}$  using $\{X_a, \tilde{X}_a\}$ and $\{X_b, \tilde{X}_a\}$, respectively. The distance from the source mode $a$ to the target $b$ is defined to be:
\begin{equation}\label{dTAS_eq}
    s[a, b] := \frac{1}{\sqrt{2}} \textbf{trace} \Big(H_{a} + H_{b} - 2H_{a}^{1/2}H_{b}^{1/2} \Big)^{{1/2}}.
\end{equation}
\end{definition}

To simplify Equation~(\ref{dTAS_eq}), we approximate the Hessian matrices with their normalized diagonals as computing the full Hessian matrices in the large parameter space of neural networks can be computationally expensive. Hence, dMAS can be expressed as follows:
\begin{equation}
    s[a, b] = \frac{1}{\sqrt{2}} \norm{H_{a}^{1/2} - H_{b}^{1/2}}_F
\end{equation}
The procedure to compute dMAS is outlined in function \texttt{dMAS()} in Algorithm~\ref{alg1}. Our metric spans a range from $0$ to $1$, where $0$ signifies a perfect match, while $1$ indicates complete dissimilarity. It is important to note that dMAS exhibits an asymmetric nature, reflecting the inherent ease of knowledge transfer from a complex model to a simpler one, as opposed to the reverse process.

\subsection{Comparison with FID}
In contrast to the statistical biases observed in metrics such as IS~\citep{salimans2016improved} and FID~\citep{heusel2017gans}~\citep{chong2020effectively}, dMAS is purposefully crafted to cater to our specific scenario of interest. It takes into account the state of the cGAN model, encompassing both the discriminator and the generator. This sets it apart from FID, which uses the GoogleNet Inception model to measure the Wasserstein distance to the ground truth distribution. Consequently, it falls short in evaluating the quality of generators and discriminators. Instead of assessing the similarity between Gaussian-like distributions, our proposed dMAS quantifies the Fisher Information distance between between the model weights. Thus, it accurately reflects the current states of the source models. Furthermore, FID has exhibited occasional inconsistency with human judgment, leading to suboptimal knowledge transfer performance~\citep{liu2018improved, wang2018transferring}. In contrast, our measure aligns more closely with human intuition and consistently demonstrates its reliability. It is important to emphasize that dMAS is not limited to the analysis of image data samples; it can be effectively applied to a wide range of data types, including text and multi-modal datasets.

\subsection{Mode-Aware Continual Learning Framework}

\begin{figure}
\centering
\includegraphics[width=0.95\textwidth]{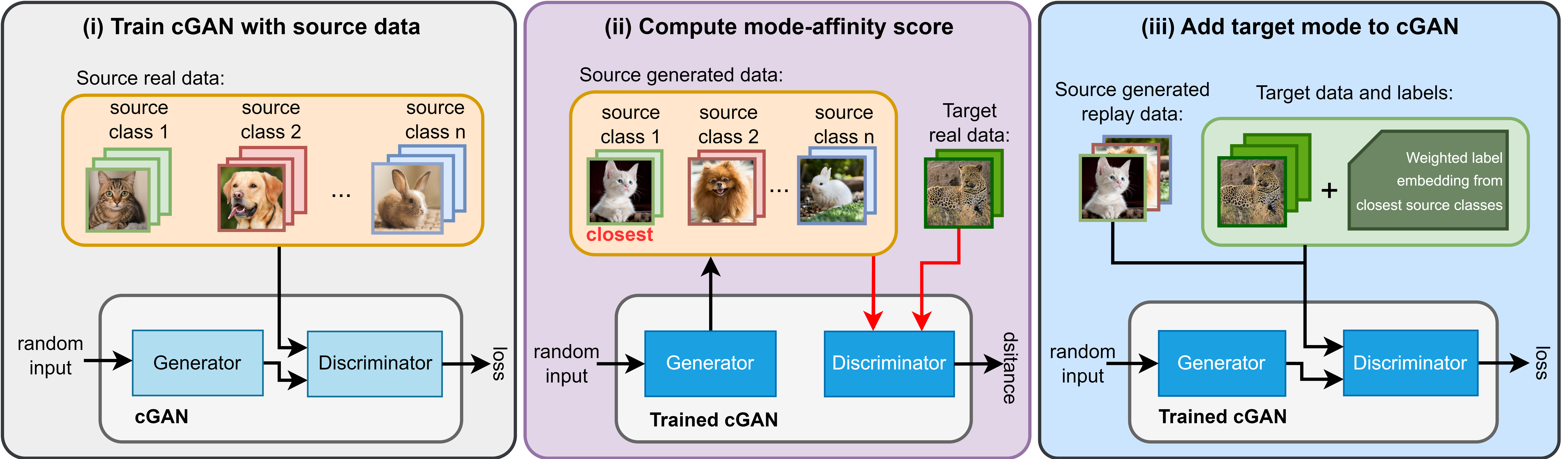}
\caption{The overview of mode-aware continual learning framework for the conditional Generative Adversarial Network: (i) Representing source data classes using cGAN, (ii) Computing the mode-affinity from each source mode to the target, (iii) Fine-tuning the generative model using the target data and the weighted label embedding from relevant modes for continual learning.}
\label{fig:continual-learning}
\end{figure}

We utilize the discriminator-based mode affinity score (dMAS) to continual learning for image generation. The goal is to train a lifelong learning cGAN model to learn new modes while avoiding catastrophic forgetting of existing modes. Consider a scenario where each generative task represents a distinct class of data. At time $t$, the cGAN model has $S$ modes corresponding to $S$ learned tasks. Here, we introduce a \textit{mode-aware continual learning} framework that allows the model to add a new mode while retaining knowledge from previous modes. We begin by embedding the numeric label of each data sample in cGAN, using an embedding layer in both the generator $\mathcal{G}$ and the discriminator $\mathcal{D}$ models. We then modify cGAN to enable it to take a linear combination of label embeddings for the target data. These label embeddings correspond to the most relevant modes, and the weights for these embedding features are associated with the computed dMAS from the related modes to the target. This enables cGAN to add a new target mode while maintaining all existing modes. Let $\texttt{emb}()$ denote the embedding layers in the generator $\mathcal{G}$ and the discriminator $\mathcal{D}$, and $C$ be the set of the relevant modes, $C=\{i_1^*, i_2^*, \ldots, i_n^*\}$. The computed mode-affinity scores from these modes to the target are denoted as $s_i^*$. Let $\sum s_i^*$ denote the total distance from all the relevant modes to the target. The label embedding for the target data samples is described as follows: 
\begin{equation}\label{label_embedding}
    \texttt{emb}(y_{target}) = \frac{s_{i_1^*}}{\sum s_i^*} \texttt{emb}(y_{train_{i_1^*}}) + \frac{s_{i_2^*}}{\sum s_i^*} \texttt{emb}(y_{train_{i_2^*}}) + ... + \frac{s_{i_n^*}}{\sum s_i^*} \texttt{emb}(y_{train_{i_n^*}})
\end{equation}

To add the new target mode without forgetting the existing learned modes, we use the target data with the above label embedding to fine-tune the cGAN model. Additionally, we utilize memory replay~\citep{robins1995catastrophic, chenshen2018memory} to prevent catastrophic forgetting. Particularly, samples generated from relevant existing modes are used to fine-tune cGAN.  The overview of the proposed approach is illustrated in Figure~\ref{fig:continual-learning}. During each iteration, training with the target data and replaying relevant existing modes are jointly implemented using an alternative optimization process. The detail of the framework is provided in Algorithm~\ref{alg1}. By applying the closest modes' labels to the target data samples in the embedding space, we can precisely update part of cGAN without sacrificing the generation performance of other existing modes. Overall, utilizing knowledge from past experience helps enhance the performance of cGAN in learning new modes while reducing the amount of the required training data samples. Next, we provide a theoretical analysis of our proposed method.

\begin{theorem}
\label{theorem1}
Let $\theta$ be the model's parameters and $X_a, X_b$ be the source and target data, with the density functions $p_a, p_b$, respectively.
Assume the loss functions $L_a(\theta) = \mathbb{E}[l(X_a;\theta)]$ and $L_b(\theta) = \mathbb{E}[l(X_b;\theta)]$ are strictly convex and have distinct global minima. 
Let $X_n$ be the mixture of $X_a$ and $X_b$, described by $p_n = \alpha p_a + (1-\alpha) p_b$, where $\alpha \in (0,1)$. The corresponding loss function is $L_n(\theta) = \mathbb{E}[l(X_n;\theta)]$.
Under these assumptions, it follows that $\theta^* = \arg \min_{\theta}L_n(\theta)$ satisfies:
\begin{equation}
    L_a(\theta^*) > \min_{\theta} L_a(\theta)
\end{equation}
\end{theorem}

In the above theorem, the introduction of a new mode through mode injection inherently involves a trade-off between the mode-adding ability and the potential performance loss compared to the original model. In essence, when incorporating a new mode, the performance of existing modes cannot be improved. The detailed proof of Theorem~\ref{theorem1} is provided in Appendix~\ref{appendix-proof}.

\begin{algorithm}[t]
\SetKwInput{KwInput}{Input}         
\SetKwInput{KwOutput}{Output}       
\SetKwInput{KwData}{Data}
\SetKwFunction{FewShot}{Main}
\SetKwFunction{DTAS}{dMAS}
\SetKwFunction{MaxMatching}{MaximumMatching}
\SetKwComment{Comment}{$\triangleright$ }{}
\SetCommentSty{scriptsize}

\DontPrintSemicolon

\KwData{Source data: $(X_{train}, y_{train})$, Target data: $X_{target}$}
\KwInput{The generator $\mathcal{G}$ and discriminator $\mathcal{D}$ of cGAN}
\KwOutput{Continual learning generator $\mathcal{G}_{\Bar{\theta}}$ and discriminator $\mathcal{D}_{\Bar{\theta}}$}

    \SetKwProg{Fn}{Function}{:}{}
    \Fn{\DTAS{$X_a, y_a, X_b, \mathcal{G}, \mathcal{D}$}}{
        Generate data $\tilde{X}_a$ of class label $y_a$ using the generator $\mathcal{G}$\;
        Compute $H_{a}$ from the loss of discriminator $\mathcal{D}$ using $\{X_a, \tilde{X}_a\}$\;
        Compute $H_{b}$ from the loss of discriminator $\mathcal{D}$ using $\{X_b, \tilde{X}_a\}$\;
        \KwRet $\displaystyle s[a, b] = \frac{1}{\sqrt{2}} \norm{H_{a}^{1/2} - H_{b}^{1/2}}_F$
    }
    \SetKwProg{Fn}{Function}{:}{} \Fn{\FewShot}{ 
        Train ($\mathcal{G}_{\theta}$, $\mathcal{D}_{\theta}$) with $X_{train}, y_{train}$  \Comment*[r]{Pre-train cGAN model}
        Construct S source modes, each from a data class in $y_{train}$\;

        \For{$i=1,2,\ldots,S$}{ 
            $s_i = \DTAS(X_{train_i}, y_{train_i}, X_{target}, \mathcal{G}_{\theta}, \mathcal{D}_{\theta})$ \Comment*[r]{Find the closest modes}
        }
        \KwRet closest mode(s): $i^* = \underset{i}{\mathrm{argmin}}\ s_i$\;

        Generate data $X_{i^*}$ of label $i^*$ from closest mode(s)  \Comment*[r]{Fine-tune for continual learning}
        Define the target label embedding as a linear combination of the label embeddings of the closest modes, where the weights corresponding to $s_{i^*}$ as follows:
        $\displaystyle \texttt{emb}(y_{target}) = 
        \frac{s_{i_1^*}}{\sum s_i^*} \texttt{emb}(y_{train_{i_1^*}}) + \ldots + \frac{s_{i_n^*}}{\sum s_i^*} \texttt{emb}(y_{train_{i_n^*}})$\;
       
        \While{$\theta$ not converged}{
            Update $\mathcal{G}_{\theta}$, $\mathcal{D}_{\theta}$ using real data $X_{target}$ and label embedding $\texttt{emb}(y_{target})$\;
            Replay $\mathcal{G}_{\theta}$, $\mathcal{D}_{\theta}$ with generated data $X_{i^*}$ and label embedding $\texttt{emb}(y_{train_{i^*}})$\;
        }
        \KwRet $\mathcal{G}_{\Bar{\theta}}, \mathcal{D}_{\Bar{\theta}}$\;

    }
\caption{Mode-Aware Continual Learning for Conditional Generative Adversarial Networks}
\label{alg1}
\end{algorithm}

\section{Experimental Study}\label{sec:experiment}
Our experiments aim to evaluate the effectiveness of the proposed mode-affinity measure in the continual learning framework, as well as the consistency of the discriminator-based mode affinity score for cGAN. We consider a scenario where each generative task corresponds to a single data class in the MNIST~\citep{lecun2010mnist}, CIFAR-10~\citep{krizhevsky2009learning}, CIFAR-100~\citep{krizhevsky2009learning}, and Oxford Flower~\citep{nilsback2008automated} datasets. Here, we compare the proposed framework with baselines and state-of-the-art approaches, including individual learning~\citep{mirza2014conditional}, sequential fine-tuning~\citep{wang2018transferring}, multi-task learning~\citep{standley2020tasks}, FID-transfer learning~\citep{wang2018transferring}, EWC-GAN~\citep{seff2017continual}, Lifelong-GAN~\citep{zhai2019lifelong}, and CAM-GAN~\citep{varshney2021cam}. The results show the efficacy of our approach in terms of generative performance and the ability to learn new modes while preserving knowledge of the existing modes.

\subsection{Mode Affinity Score Consistency}
\begin{figure}
\centering
    \begin{subfigure}{0.32\textwidth}
    \centering
    \includegraphics[width=0.92\textwidth]{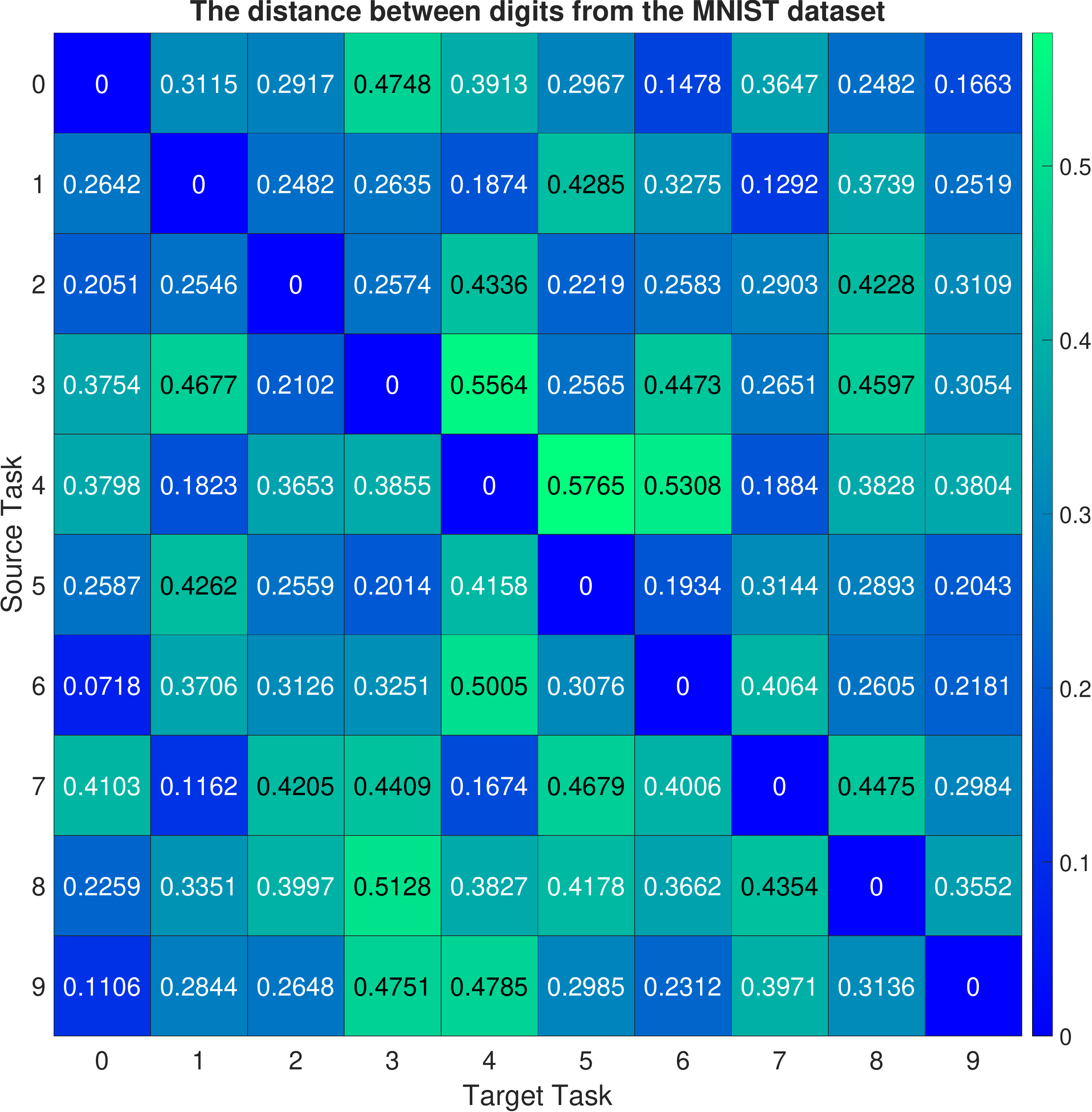}
    \caption{MNIST}
    \end{subfigure}
    \begin{subfigure}{0.32\textwidth}
    \centering
    \includegraphics[width=\textwidth]{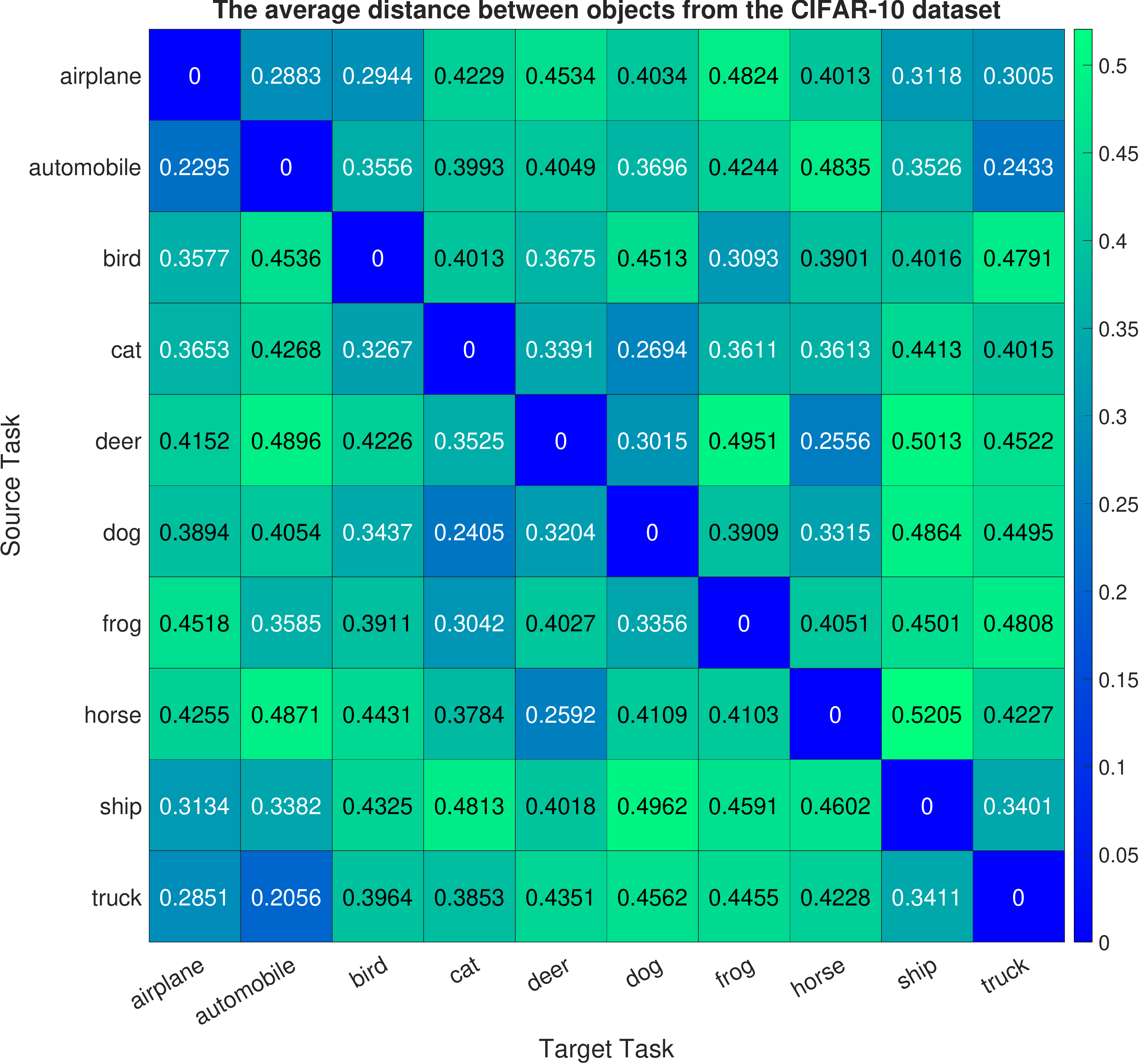}
    \caption{CIFAR-10}
    \end{subfigure}
    \begin{subfigure}{0.32\textwidth}
    \centering
    \includegraphics[width=0.96\textwidth]{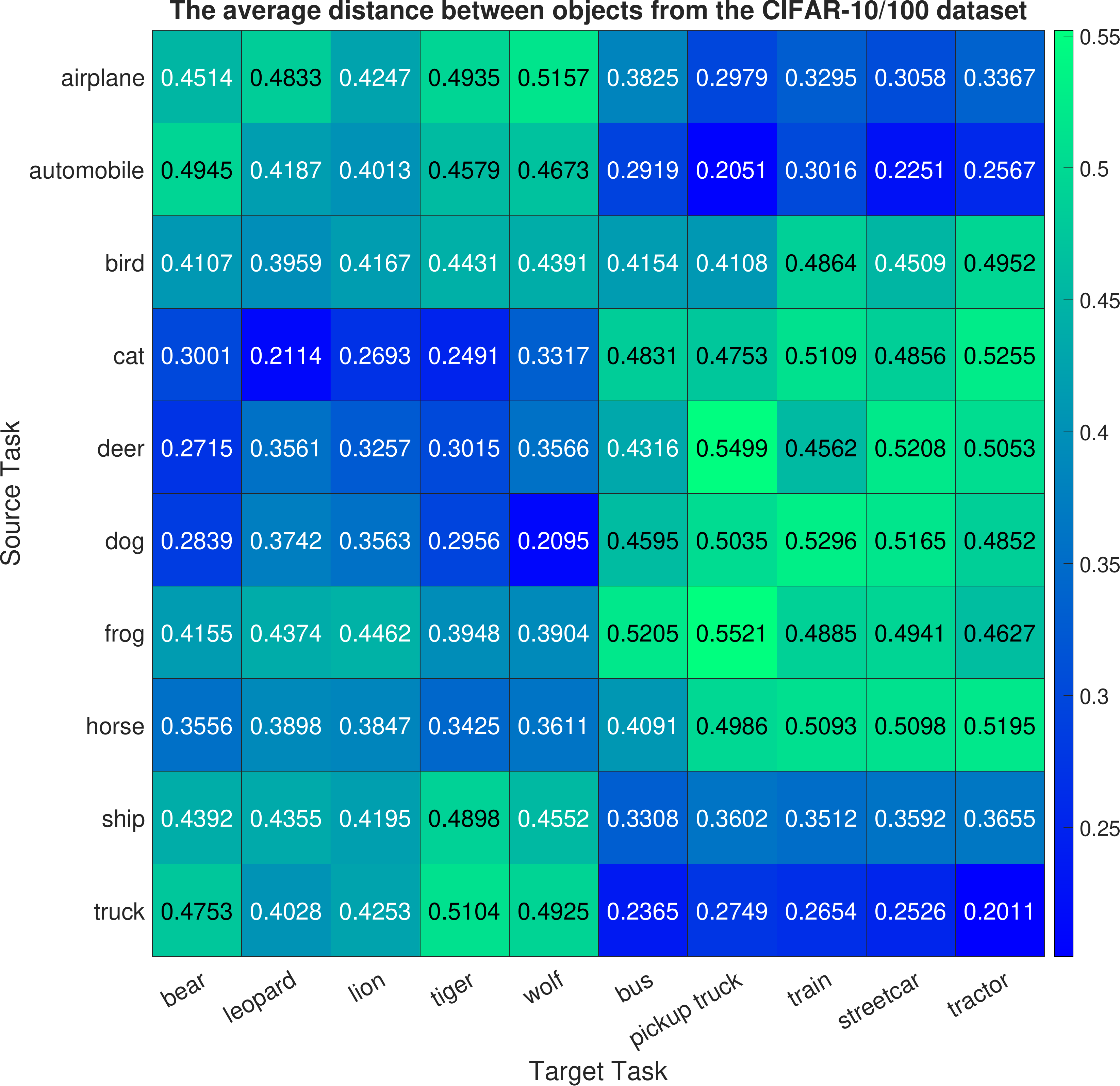}
    \caption{CIFAR-100}
    \end{subfigure}
\caption{The mean values of the mode-affinity scores computed across 10 trial runs using the cGAN model for data classes from (a) MNIST, (b) CIFAR-10, (c) CIFAR-100 datasets.}
\label{fig:mean-distance}
\end{figure}

\begin{figure}
\centering
    \begin{subfigure}{0.32\textwidth}
    \centering
    \includegraphics[width=\textwidth]{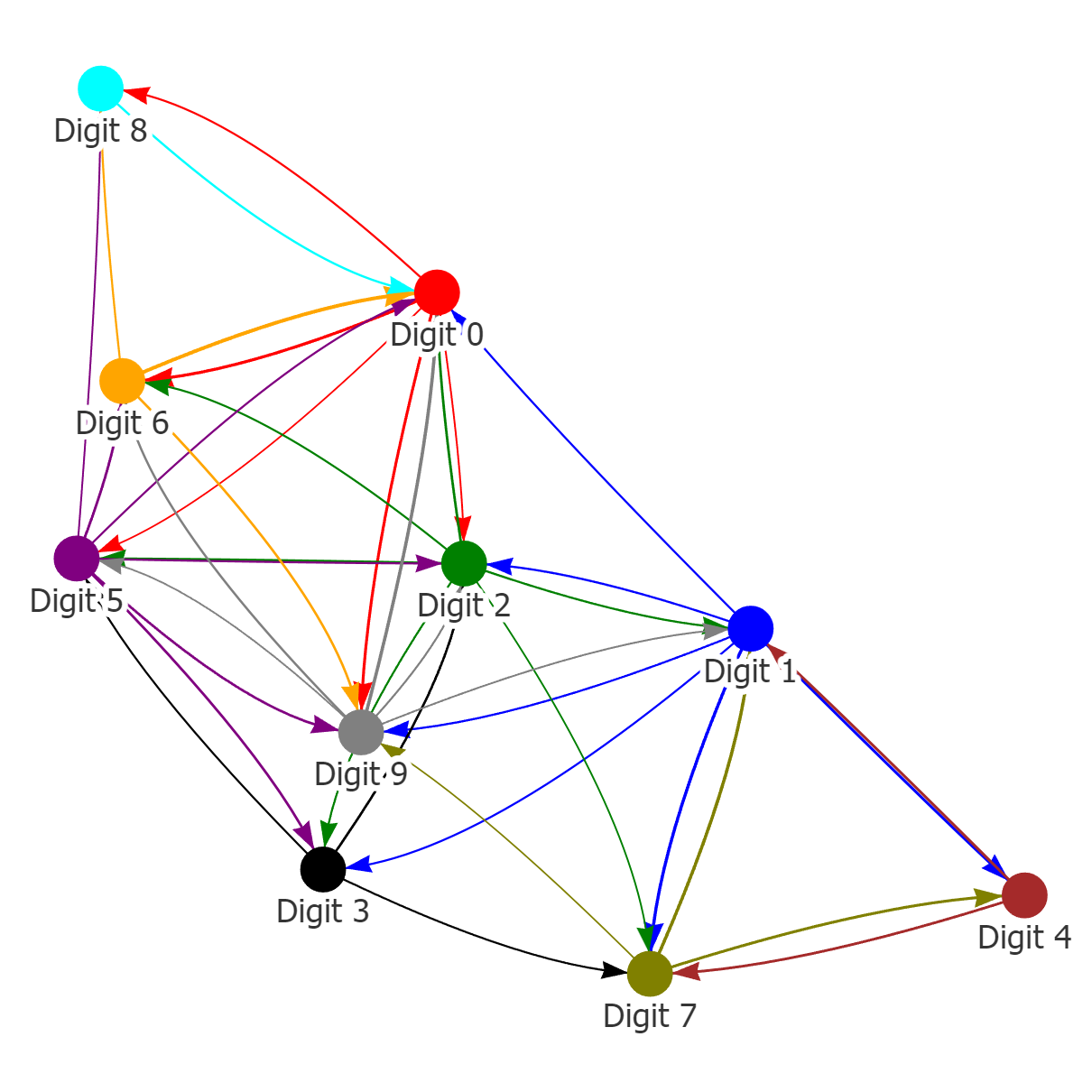}
    \caption{MNIST digits}
    \end{subfigure}
    \begin{subfigure}{0.32\textwidth}
    \centering
    \includegraphics[width=\textwidth]{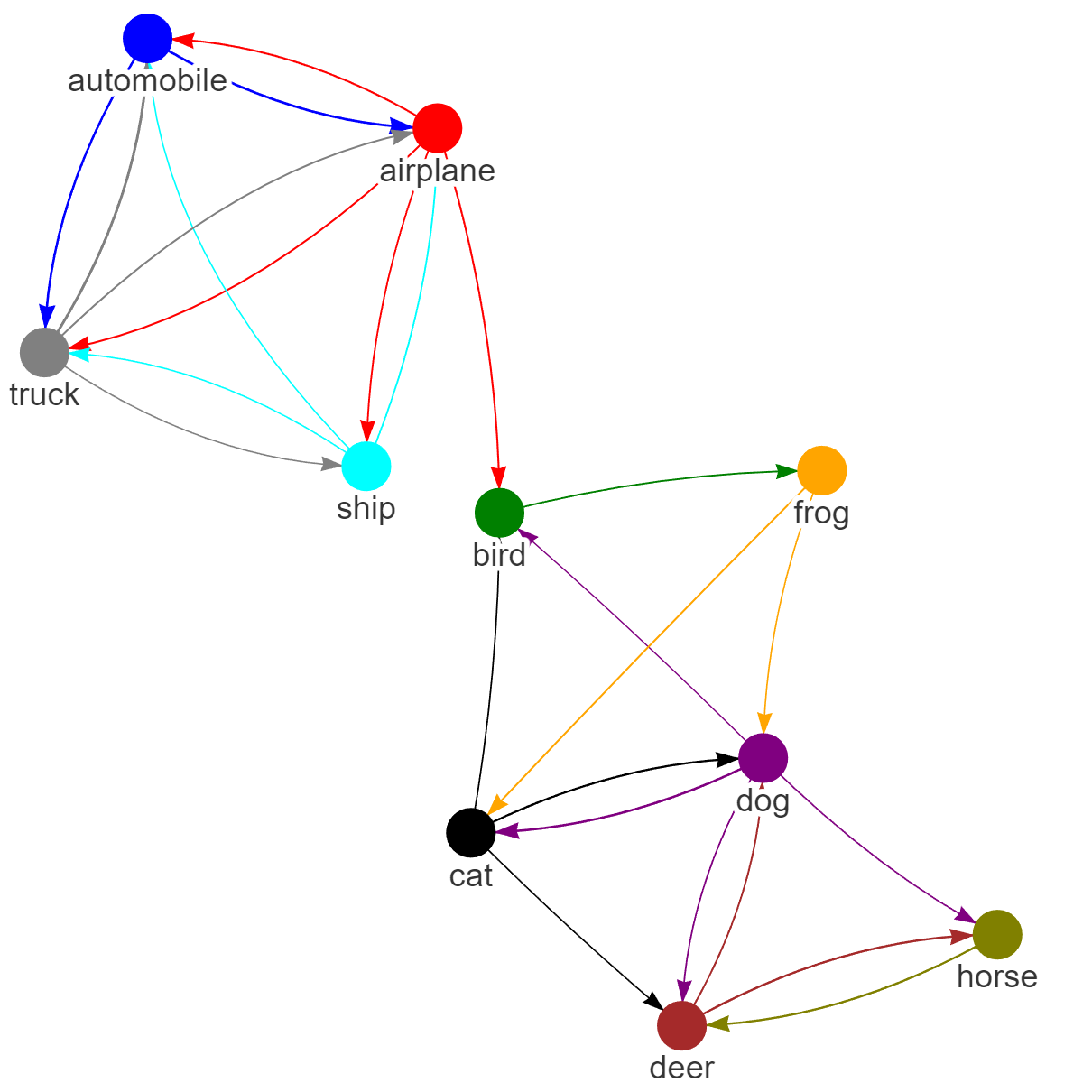}
    \caption{CIFAR-10 objects}
    \end{subfigure}
    \begin{subfigure}{0.32\textwidth}
    \centering
    \includegraphics[width=\textwidth]{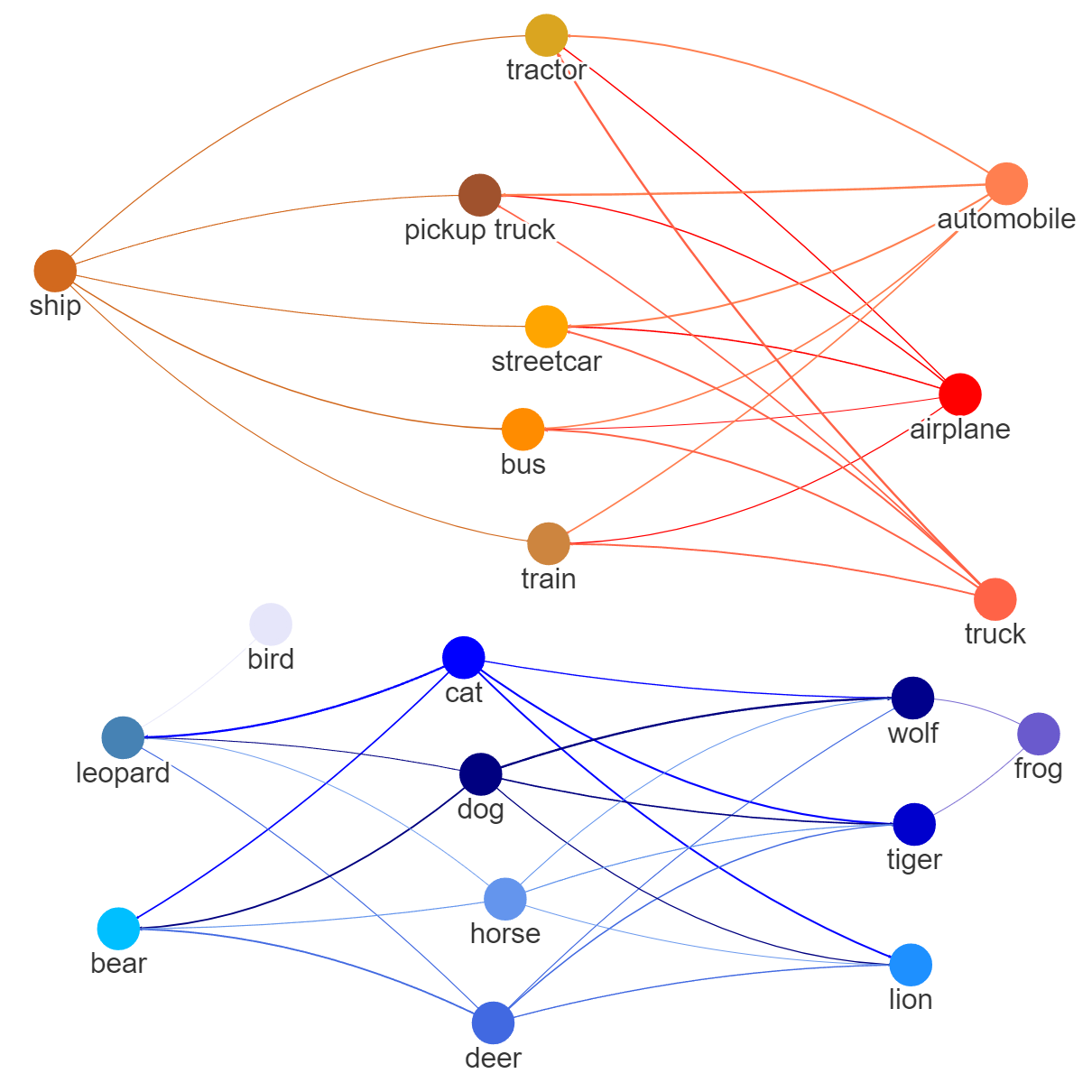}
    \caption{CIFAR-100 objects}
    \end{subfigure}
\caption{The atlas plot from the computed distances using the cGAN model for data classes from (a) MNIST, (b) CIFAR-10, (c) CIFAR-100 datasets.}
\label{fig:atlas-distance}
\end{figure}

In the first experiment, the ten generative tasks are defined based on the MNIST dataset, where each task corresponds to generating a specific digit (i.e., $0, 1, \ldots, 9$). For instance, task 0 aimed at generating images representing digit 0, while task 1 aimed at generating images depicting digit 1. The cGAN model was trained to generate images from the nine source tasks while considering one task as the target task. The cGAN model has nine modes corresponding to nine source tasks. The well-trained generator of this cGAN model served as the representation network for the source data. To evaluate the consistency of the closest modes for each target, we conduct $10$ trial runs, in which the source cGAN model is initialized randomly. The mean and standard deviation of the mode-affinity scores between each pair of source-target modes are shown in Figure~\ref{fig:mean-distance} (a) and Figure~\ref{fig:mnist-distance-var}, respectively. In the mean table from Figure~\ref{fig:mean-distance} (a), the columns denote the distance from each source mode to the given target. For instance, the first column indicates that digits $6$ and $9$ are closely related to the target digit $0$. Similarly, the second column shows that digits $4$ and $7$ are the closest tasks to the incoming digit $1$. The standard deviation table from Figure~\ref{fig:mnist-distance-var} indicates that the calculated distance is stable, as there are no overlapping fluctuations and the orders of similarity between tasks are preserved across $10$ runs. In other words, this suggests that the tendency of the closest modes for each target remains consistent regardless of the initialization of cGAN. Thus, the computed mode-affinity score demonstrates consistent results. Additionally, we provide the atlas plot in Figure~\ref{fig:atlas-distance}(a) which gives an overview of the relationship between the digits based on the computed distances. The plot reveals that digits $1, 4, 7$ exhibit a notable similarity, while digits $0, 6, 8$ are closely related. This plot provides a useful visualization to showcase the pattern and similarity among different digits.

Analogously, we define ten generative tasks for the CIFAR-10 dataset, each corresponding to a specific object, such as airplane, automobile, bird, cat, deer, dog, frog, horse, ship, and truck. Following the previous experiment, one task is designated as the target task, while the others are the source tasks used to train the cGAN model for image generation. The generator of the well-trained cGAN model serves as the representation network for the source tasks. We present the mean and standard deviation of computed mode-affinity scores between each pair of source-target modes over $10$ trial runs in Figure~\ref{fig:mean-distance}(b) and Figure~\ref{fig:cifar-distance-var}, respectively. The mean table in Figure~\ref{fig:mean-distance}(b) shows the average distance of each source mode from the target (e.g., trucks are similar to automobiles, and cats are closely related to dogs). The standard deviation table in Figure~\ref{fig:cifar-distance-var} demonstrates the stability of the results across different initialization of cGAN. The consistent findings suggest that the computed distances from the CIFAR-10 dataset are reliable. Furthermore, in Figure~\ref{fig:atlas-distance}(b), we include the atlas plot which presents an overview of the relationship between the objects based on the computed mode-affinity scores. The plot reveals that automobile, truck, ship, and airplane have a strong connection, while the other classes (i.e., bird, cat, dog, deer, horse, frog) also exhibit a significant resemblance. 

Next, the CIFAR-100 dataset is utilized to define ten target tasks, each corresponding to a specific image class, such as bear, leopard, lion, tiger, wolf, bus, pickup truck, train, streetcar, and tractor. For this experiment, the cGAN model is trained on the entire CIFAR-10 dataset to generate images from ten classes or modes. Figure~\ref{fig:mean-distance}(c) and Figure~\ref{fig:cifar100-distance-var} respectively display the mean and standard deviation of the computed mode-affinity scores between the source-target modes. The mean table in Figure~\ref{fig:mean-distance}(c) indicates the average distance from each CIFAR-10 source mode to the CIFAR-100 target mode. Notably, the target tasks of generating bear, leopard, lion, tiger, and wolf images are closely related to the group of cat, deer, and dog. Specifically, cat images are closely related to leopard, lion, and tiger images. Moreover, the target modes of generating bus, pickup truck, streetcar, and tractor images are highly related to the group of automobile, truck, airplane, and ship. The standard deviation table in Figure~\ref{fig:cifar100-distance-var} also indicates that the computed distances are consistent across different trial runs, demonstrating the stability of the computed mode-affinity scores. Thus, the distances computed between the CIFAR-100 modes and the CIFAR-10 modes are reliable and consistent. In addition, Figure~\ref{fig:atlas-distance}(c) includes an atlas plot that provides a visual representation of the relationships between objects based on the computed distances. The plot reveals a strong connection between the vehicles, such as tractors, trucks, and trains, as well as a notable closeness between the animal classes, such as lions, tigers, cats, and dogs. This plot serves as a useful tool for visualizing the similarity and relationship between different objects in the CIFAR-10 and CIFAR-100 datasets and can help to identify relevant data classes for the target class.

\begin{table}[t]
\caption{Knowledge transfer performance of mode-affinity score against other baselines and FID transfer learning approaches for 10-shot, 20-shot, and 100-shot.}
\label{table:fid-compare}
\begin{center}
\begin{tabular}{l|c|ccc}
\hline
\multicolumn{1}{l}{\textbf{Approach}} &\multicolumn{1}{c}{\textbf{Target}} &\multicolumn{1}{c}{\textbf{10-shot}} &\multicolumn{1}{c}{\textbf{20-shot}} &\multicolumn{1}{c}{\textbf{100-shot}} \\
\hline
Individual Learning~\citep{mirza2014conditional}           
& Bus  & 94.82 & 89.01  & 78.47 \\
Sequential Fine-tuning~\citep{zhai2019lifelong}           
& Bus  & 88.03 & 79.51 & 67.33 \\
Multi-task Learning~\citep{standley2020tasks}             
& Bus  & 80.06 & 76.33  & 61.59 \\
FID-Transfer Learning~\citep{wang2018transferring}             
& Bus  & 61.34 & 54.18  & 46.37 \\
\hline
\textbf{MA-Transfer Learning (ours)}      
& \textbf{Bus}  & \textbf{57.16} & \textbf{50.06}  & \textbf{41.81} \\
\hline
\end{tabular}
\end{center}
\end{table}

Additionally, we conducted a knowledge transfer experiment to assess the effectiveness of the proposed mode affinity score in transfer learning scenarios utilizing the MNIST, CIFAR-10, and CIFAR-100 datasets. In these experiments, we designated one data class as the target, while the remaining nine classes served as source tasks. Our method first computes the dMAS distance from the target to each source task. After identifying the closest task, we fine-tune the cGAN model using the target data samples with the label from the closest task. This approach helps the cGAN model to update the specific part of the model to efficiently and quickly learn the target task. The transfer learning framework is illustrated in Figure~\ref{fig:transfer-learning} and the pseudo-code is provided in Algorithm~\ref{alg2}. The image generation performance in terms of FID scores are presented in Tables~\ref{table:fid-compare} and ~\ref{transfer-learning-table}. Notably, our utilization of dMAS for knowledge transfer consistently outperforms the baseline methods, including Individual Learning, Sequential Fine-tuning, and Multi-task Learning. Remarkably, our approach achieves superior results while utilizing only 10\% of the target data samples. When comparing our method with the FID transfer learning approach~\citep{wang2018transferring}, we observe similar performance in most scenarios. However, in the CIFAR-100 experiment, where the target class is bus images, we intentionally trained the source model using only a limited number of truck samples, leading to an inadequately trained source model. Consequently, FID still regards the class of truck images as the closest to the target, resulting in less efficient knowledge transfer. In contrast, our method takes into account the state of the models and selects the task closest to automobile images, resulting in more effective knowledge transfer. Table~\ref{table:fid-compare} clearly demonstrates that our approach outperforms FID transfer learning in 10-shot, 20-shot, and 100-shot scenarios.

\begin{table}[t]
\caption{Comparison of the mode-aware continual learning framework for cGAN against other baseline and state-of-the-art approaches for MNIST, CIFAR-10, and CIFAR-100, in terms of FID.}
\label{continual-learning-table}
\begin{center}
\begin{tabular}{l|c|ccc}
\hline
\multicolumn{1}{l}{} &\multicolumn{1}{c}{} &\multicolumn{3}{c}{\textbf{MNIST}} \\
\multicolumn{1}{l}{\textbf{Approach}} &\multicolumn{1}{c}{\textbf{Target}} &\multicolumn{1}{c}{\boldmath{$\mathcal{P}_{target}$}} &\multicolumn{1}{c}{\boldmath{$\mathcal{P}_{closest}$}} &\multicolumn{1}{c}{\boldmath{$\mathcal{P}_{average}$}} \\
\hline
Individual Learning~\citep{mirza2014conditional}           
& Digit 0  & 19.62 & - & - \\
Sequential Fine-tuning~\citep{zhai2019lifelong}           
& Digit 0  & 16.72 & 26.53 & 26.24 \\
Multi-task Learning~\citep{standley2020tasks}             
& Digit 0  & 11.45 & \textbf{5.83} & 6.92 \\
EWC-GAN~\citep{seff2017continual}
& Digit 0  & 8.96 & 7.51 & 7.88 \\
Lifelong-GAN~\citep{zhai2019lifelong}           
& Digit 0  & 8.65 & 6.89 & 7.37 \\
CAM-GAN~\citep{varshney2021cam}           
& Digit 0  & 7.02 & 6.43 & 6.41 \\
\hline
\textbf{MA-Continual Learning (ours)}     & \textbf{Digit 0}  & \textbf{6.32} & 5.93  & \textbf{5.72} \\

\hline
Individual Learning~\citep{mirza2014conditional}           
& Digit 1  & 20.83 & - & - \\
Sequential Fine-tuning~\citep{zhai2019lifelong}           
& Digit 1  & 18.24 & 26.73 & 27.07 \\
Multi-task Learning~\citep{standley2020tasks}             
& Digit 1  & 11.73 & 6.51 & 6.11 \\
EWC-GAN~\citep{seff2017continual}
& Digit 1  & 9.62 & 8.65 & 8.23 \\
Lifelong-GAN~\citep{zhai2019lifelong}           
& Digit 1  & 8.74 & 7.31 & 7.29 \\
CAM-GAN~\citep{varshney2021cam}           
& Digit 1  & 7.42 & 6.58 & 6.43 \\
\hline
\textbf{MA-Continual Learning (ours)}     & \textbf{Digit 1}  & \textbf{6.45} & \textbf{6.14} & \textbf{5.92} \\

\hline

\multicolumn{1}{l}{} &\multicolumn{1}{c}{} &\multicolumn{3}{c}{\textbf{CIFAR-10}} \\
\multicolumn{1}{l}{\textbf{Approach}} &\multicolumn{1}{c}{\textbf{Target}} &\multicolumn{1}{c}{\boldmath{$\mathcal{P}_{target}$}} &\multicolumn{1}{c}{\boldmath{$\mathcal{P}_{closest}$}} &\multicolumn{1}{c}{\boldmath{$\mathcal{P}_{average}$}} \\
\hline
Individual Learning~\citep{mirza2014conditional}           
& Truck  & 72.18 & - & - \\
Sequential Fine-tuning~\citep{zhai2019lifelong}           
& Truck  & 61.52 & 65.18 & 64.62 \\
Multi-task Learning~\citep{standley2020tasks}             
& Truck  & 55.32 & \textbf{33.65}  & 35.52 \\
EWC-GAN~\citep{seff2017continual}
& Truck  & 44.61 & 35.54 & 35.21 \\
Lifelong-GAN~\citep{zhai2019lifelong}           
& Truck  & 41.84 & 35.12 & 34.67 \\
CAM-GAN~\citep{varshney2021cam}           
& Truck  & 37.41 & 34.67 & 34.24 \\
\hline
\textbf{MA-Continual Learning (ours)}     & \textbf{Truck}  & \textbf{35.57} & 34.68  & \textbf{33.89} \\

\hline
Individual Learning~\citep{mirza2014conditional}           
& Cat  & 65.18 & - & - \\
Sequential Fine-tuning~\citep{zhai2019lifelong}           
& Cat  & 61.36 & 67.82 & 65.23 \\
Multi-task Learning~\citep{standley2020tasks}             
& Cat  & 54.47 & \textbf{34.55}  & 36.74 \\
EWC-GAN~\citep{seff2017continual}
& Cat  & 45.17 & 36.53 & 35.62 \\
Lifelong-GAN~\citep{zhai2019lifelong}           
& Cat  & 42.58 & 35.76 & 34.89 \\
CAM-GAN~\citep{varshney2021cam}           
& Cat  & 37.29 & 35.28 & 34.62 \\
\hline
\textbf{MA-Continual Learning (ours)}     & \textbf{Cat}  & \textbf{35.29} & 34.76  & \textbf{34.01} \\

\hline

\multicolumn{1}{l}{} &\multicolumn{1}{c}{} &\multicolumn{3}{c}{\textbf{CIFAR-100}} \\
\multicolumn{1}{l}{\textbf{Approach}} &\multicolumn{1}{c}{\textbf{Target}} &\multicolumn{1}{c}{\boldmath{$\mathcal{P}_{target}$}} &\multicolumn{1}{c}{\boldmath{$\mathcal{P}_{closest}$}} &\multicolumn{1}{c}{\boldmath{$\mathcal{P}_{average}$}} \\
\hline
Individual Learning~\citep{mirza2014conditional}           
& Lion  & 72.58 & - & - \\
Sequential Fine-tuning~\citep{zhai2019lifelong}           
& Lion  & 63.78 & 66.56 & 65.82 \\
Multi-task Learning~\citep{standley2020tasks}             
& Lion  & 56.32 & \textbf{36.38}  & 37.47 \\
EWC-GAN~\citep{seff2017continual}
& Lion  & 46.53 & 38.79 & 36.72 \\
Lifelong-GAN~\citep{zhai2019lifelong}           
& Lion  & 43.57 & 38.35 & 36.53 \\
CAM-GAN~\citep{varshney2021cam}           
& Lion  & 40.24 & 37.64 & 36.86 \\
\hline
\textbf{MA-Continual Learning (ours)}     & \textbf{Lion}  & \textbf{38.73} & 36.53  & \textbf{35.88} \\

\hline
Individual Learning~\citep{mirza2014conditional}           
& Bus  & 78.47 & - & - \\
Sequential Fine-tuning~\citep{zhai2019lifelong}           
& Bus  & 67.51 & 70.77 & 69.26 \\
Multi-task Learning~\citep{standley2020tasks}             
& Bus  & 61.86 & \textbf{37.21}  & 38.21 \\
EWC-GAN~\citep{seff2017continual}
& Bus  & 49.86 & 39.84 & 37.91 \\
Lifelong-GAN~\citep{zhai2019lifelong}           
& Bus  & 43.73 & 39.75 & 37.66 \\
CAM-GAN~\citep{varshney2021cam}           
& Bus  & 42.81 & 38.82 & 37.21 \\
\hline
\textbf{MA-Continual Learning (ours)}     & \textbf{Bus}  & \textbf{41.68} & 38.63  & \textbf{36.87} \\

\hline
\end{tabular}
\end{center}
\end{table}

\subsection{Continual Learning Performance}
We apply the computed mode-affinity scores between generative tasks in the MNIST, CIFAR-10, and CIFAR-100 datasets to the mode-aware continual learning framework. In each dataset, we define two target tasks for continual learning scenarios and consider the remaining eight classes as source tasks. Particularly, (digit 0, digit 1), (truck, cat), and (lion, bus) are the targets for the MNIST, CIFAR-10, and CIFAR-100 experiments, respectively. The cGAN model is trained to sequentially update these target tasks. Here, we select the top-2 closest modes to each target and leverage their knowledge for quick adaptation in learning the target task while preventing catastrophic forgetting. First, we construct a label embedding for the target data samples based on the label embeddings of the top-2 closest modes and the computed distances, as shown in Equation~(\ref{label_embedding}). Next, we fine-tune the source cGAN model with the newly-labeled target samples, while also implementing memory replay to avoid catastrophic forgetting of the existing modes. After adding the first target to cGAN, we continue the continual learning process for the second target in each experiment. We compare our framework with sequential fine-tuning~\citep{zhai2019lifelong}, multi-task learning~\citep{standley2020tasks}, EWC-GAN~\citep{seff2017continual}, lifelong-GAN~\citep{zhai2019lifelong}, and CAM-GAN~\citep{varshney2021cam} for the few-shot generative task with 100 target data samples. We report the FID scores of the images from the target mode, top-2 closest modes, and the average of all modes in Table~\ref{continual-learning-table}. By selectively choosing and utilizing the relevant knowledge from learned modes, our approach significantly outperforms the conventional training methods (i.e., sequential fine-tuning, and multi-task learning) for both the first (i.e., digit $0$, truck, lion) and the second generative tasks (i.e., digit $1$, cat, and bus). The results further demonstrate that our proposed mode-aware continual learning approach significantly outperforms EWC-GAN~\citep{seff2017continual} in the second target task in all datasets. Moreover, our model also achieves highly competitive results in comparison to lifelong-GAN~\citep{zhai2019lifelong} and CAM-GAN~\citep{varshney2021cam}, showcasing its outstanding performance on the first and second target tasks. Although we observed a slight degradation in the performance of the top-2 closest modes due to the trade-off discussed in Theorem~\ref{theorem1}, our lifelong learning model demonstrates better overall performance when considering all the learned modes.

\begin{figure}
\centering
    \begin{subfigure}{0.32\textwidth}
    \centering
    \includegraphics[width=\textwidth]{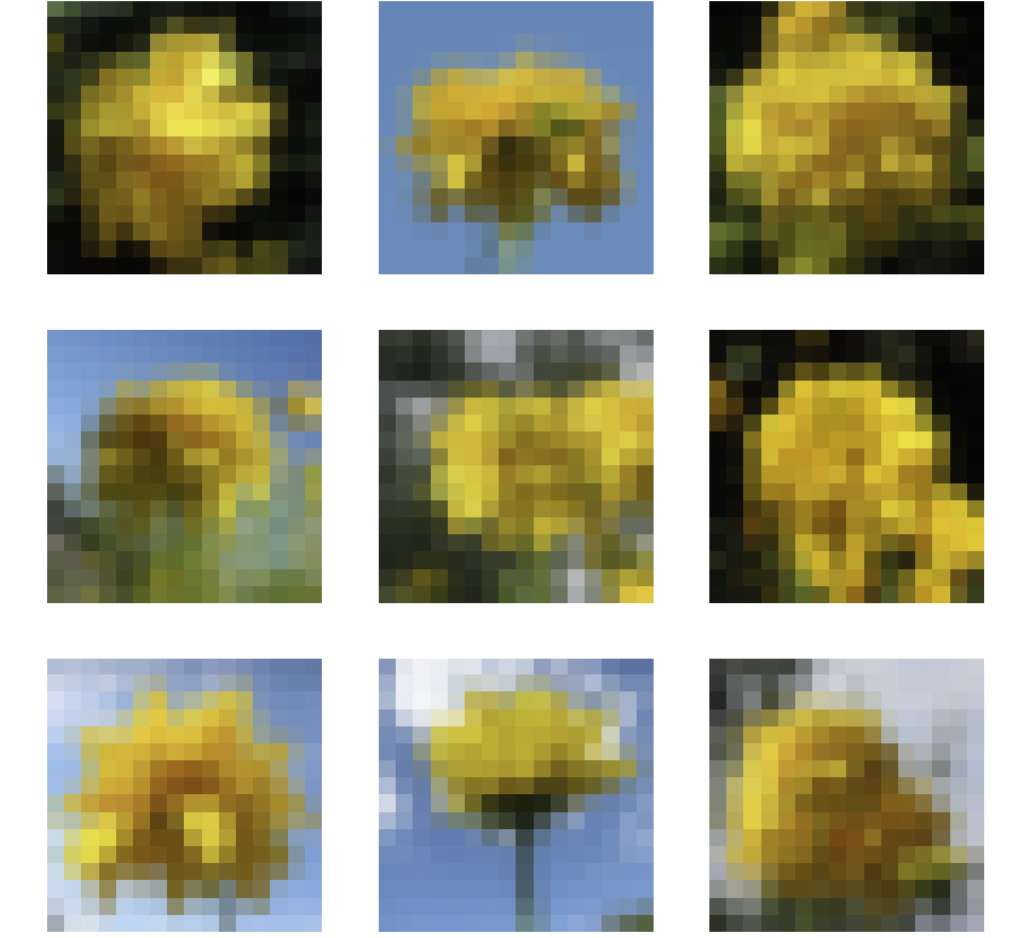}
    \caption{Goldquelle}
    \end{subfigure}
    \begin{subfigure}{0.32\textwidth}
    \centering
    \includegraphics[width=\textwidth]{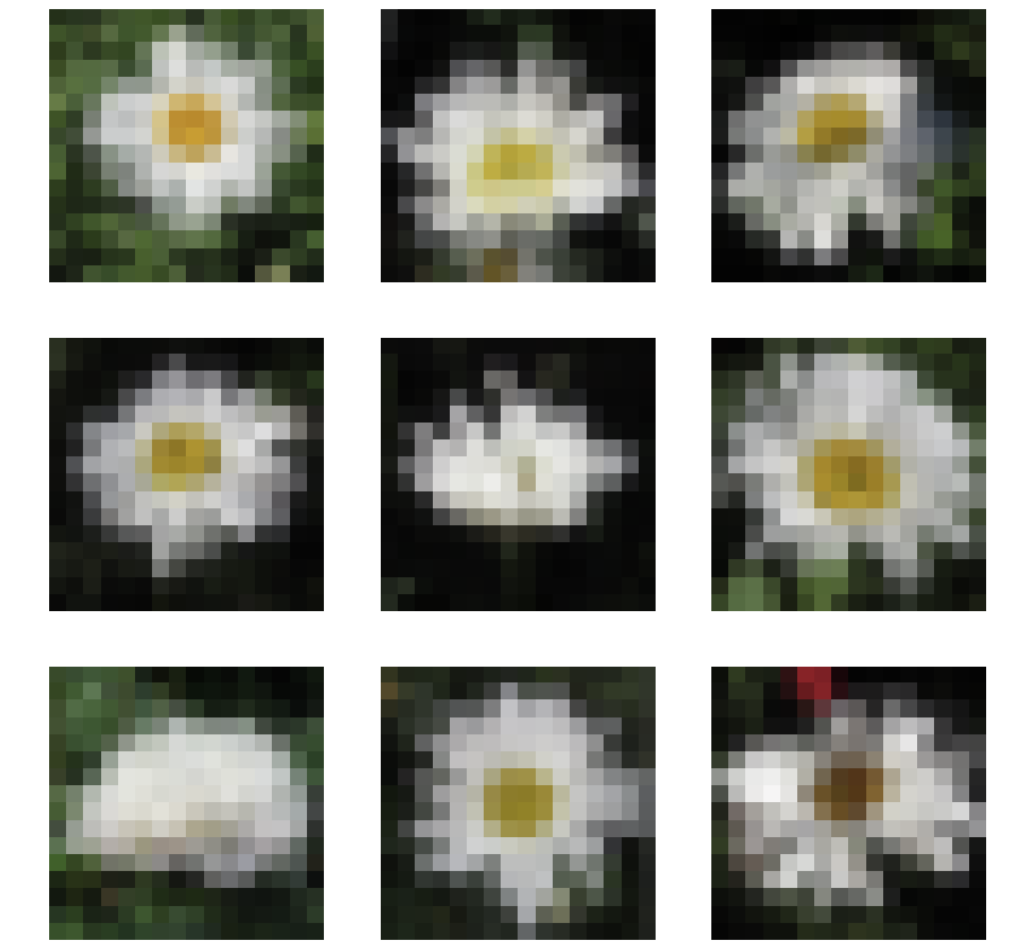}
    \caption{Shasta Daisy}
    \end{subfigure}
    \begin{subfigure}{0.32\textwidth}
    \centering
    \includegraphics[width=\textwidth]{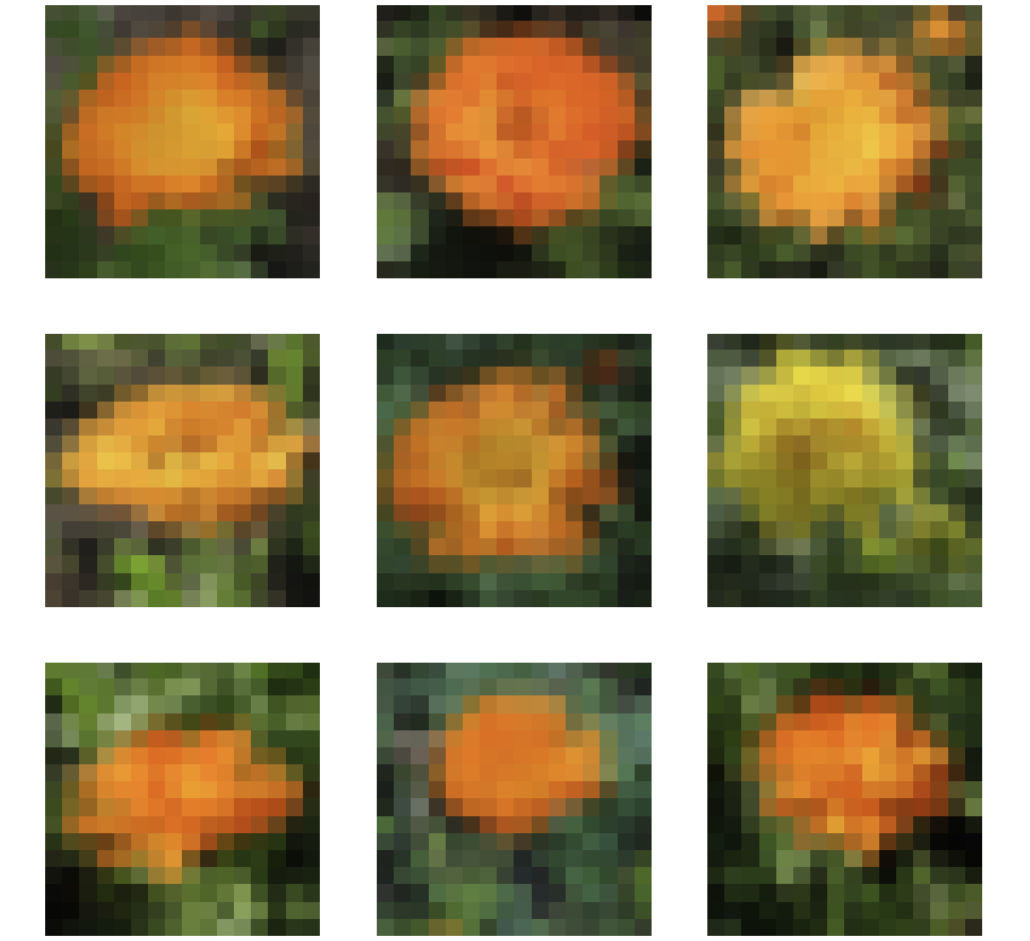}
    \caption{Calendula}
    \end{subfigure}
\caption{The generated image samples from the continual learning cGAN in Oxford Flower dataset, with top-2 relevant modes: (a) Goldquelle, (b) Shasta Daisy, and the target mode: (c) Calendula.}
\label{fig:flower}
\end{figure}

Moreover, we implement the proposed continual learning framework on the Oxford Flower dataset, specifically focusing on a subset containing ten distinct flower categories. Here, we define ten generative tasks, each corresponding to one of these flower categories. Our objective is to generate images of calendula flowers, thus designating it as the target task. The remaining flower categories, meanwhile, serve as the source tasks for training cGAN. To determine which source tasks share the greatest resemblance to the target task, we employ dMAS to compute the proximity between the target task and each of the source tasks. This analysis has unveiled the two closest tasks to calendula, namely goldquelle and shasta daisy. Hence, we leverage the knowledge from these related tasks to formulate the weighted target label. This label is subsequently employed with calendula samples to fine-tune the cGAN.
Figure~\ref{fig:flower} represents the generated images of the top-2 closest tasks (i.e., goldquelle and shasta daisy), alongside the target task (i.e., calendula), following the fine-tuning process. The results indicate that cGAN effectively leverages the inherent similarity between goldquelle and shasta daisy to enhance its ability to generate calendula flowers. However, it's worth noting that in some instances, the model may generate goldquelle-like images instead of calendula flowers. This occurrence can be attributed to the remarkably close resemblance between these two types of flowers.

\section{Conclusion}
We present a new measure of similarity between generative tasks for conditional generative adversarial networks. Our distance, called the discriminator-based mode affinity score, is based on the expectation of the Hessian matrices derived from the discriminator's loss function. This measure provides insight into the difficulty of extracting valuable knowledge from existing modes to learn new tasks. We apply this metric within the framework of continual learning, capitalizing on the knowledge acquired from relevant learned modes to expedite adaptation to new target modes. Through a series of experiments, we empirically validate the efficacy of our approach, highlighting its advantages over traditional fine-tuning methods and other state-of-the-art continual learning techniques.

\newpage
\subsubsection*{Acknowledgments}
This work was supported in part by the Army Research Office grant No. W911NF-15-1-0479.

\bibliography{iclr2024_conference}

\begin{thebibliography}{71}
\providecommand{\natexlab}[1]{#1}
\providecommand{\url}[1]{\texttt{#1}}
\expandafter\ifx\csname urlstyle\endcsname\relax
  \providecommand{\doi}[1]{doi: #1}\else
  \providecommand{\doi}{doi: \begingroup \urlstyle{rm}\Url}\fi

\bibitem[Achille et~al.(2018)Achille, Mbeng, and Soatto]{achilledynamic}
Alessandro Achille, Glen~Bigan Mbeng, and Stefano Soatto.
\newblock The dynamic distance between learning tasks: From kolmogorov complexity to transfer learning via quantum physics and the information bottleneck of the weights of deep networks.
\newblock \emph{NeurIPS Workshop on Integration of Deep Learning Theories}, 2018.

\bibitem[{Achille} et~al.(2019){Achille}, {Lam}, {Tewari}, {Ravichandran}, {Maji}, {Fowlkes}, {Soatto}, and {Perona}]{achille2019task2vec}
Alessandro {Achille}, Michael {Lam}, Rahul {Tewari}, Avinash {Ravichandran}, Subhransu {Maji}, Charless {Fowlkes}, Stefano {Soatto}, and Pietro {Perona}.
\newblock {Task2Vec: Task Embedding for Meta-Learning}.
\newblock \emph{arXiv e-prints}, art. arXiv:1902.03545, Feb. 2019.

\bibitem[Aloui et~al.(2022)Aloui, Dong, Le, and Tarokh]{aloui2022causal}
Ahmed Aloui, Juncheng Dong, Cat~P Le, and Vahid Tarokh.
\newblock Causal knowledge transfer from task affinity.
\newblock \emph{arXiv preprint arXiv:2210.00380}, 2022.

\bibitem[Arjovsky et~al.(2017)Arjovsky, Chintala, and Bottou]{arjovsky2017wasserstein}
Martin Arjovsky, Soumith Chintala, and L{\'e}on Bottou.
\newblock Wasserstein generative adversarial networks.
\newblock In \emph{International conference on machine learning}, pp.\  214--223. PMLR, 2017.

\bibitem[Azizi et~al.(2021)Azizi, Mustafa, Ryan, Beaver, Freyberg, Deaton, Loh, Karthikesalingam, Kornblith, Chen, et~al.]{azizi2021big}
Shekoofeh Azizi, Basil Mustafa, Fiona Ryan, Zachary Beaver, Jan Freyberg, Jonathan Deaton, Aaron Loh, Alan Karthikesalingam, Simon Kornblith, Ting Chen, et~al.
\newblock Big self-supervised models advance medical image classification.
\newblock In \emph{Proceedings of the IEEE/CVF International Conference on Computer Vision}, pp.\  3478--3488, 2021.

\bibitem[Brown et~al.(2020)Brown, Mann, Ryder, Subbiah, Kaplan, Dhariwal, Neelakantan, Shyam, Sastry, Askell, et~al.]{brown2020language}
Tom Brown, Benjamin Mann, Nick Ryder, Melanie Subbiah, Jared~D Kaplan, Prafulla Dhariwal, Arvind Neelakantan, Pranav Shyam, Girish Sastry, Amanda Askell, et~al.
\newblock Language models are few-shot learners.
\newblock \emph{Advances in neural information processing systems}, 33:\penalty0 1877--1901, 2020.

\bibitem[Carpenter \& Grossberg(1987)Carpenter and Grossberg]{carpenter1987massively}
Gail~A Carpenter and Stephen Grossberg.
\newblock A massively parallel architecture for a self-organizing neural pattern recognition machine.
\newblock \emph{Computer vision, graphics, and image processing}, 37\penalty0 (1):\penalty0 54--115, 1987.

\bibitem[Chen et~al.(2018)Chen, Zhang, and Dong]{chen2018coupled}
Shixing Chen, Caojin Zhang, and Ming Dong.
\newblock Coupled end-to-end transfer learning with generalized fisher information.
\newblock In \emph{Proceedings of the IEEE Conference on Computer Vision and Pattern Recognition}, pp.\  4329--4338, 2018.

\bibitem[Chenshen et~al.(2018)Chenshen, HERRANZ, Xialei, et~al.]{chenshen2018memory}
WU~Chenshen, L~HERRANZ, LIU Xialei, et~al.
\newblock Memory replay gans: Learning to generate images from new categories without forgetting [c].
\newblock In \emph{The 32nd International Conference on Neural Information Processing Systems, Montr{\'e}al, Canada}, pp.\  5966--5976, 2018.

\bibitem[Chong \& Forsyth(2020)Chong and Forsyth]{chong2020effectively}
Min~Jin Chong and David Forsyth.
\newblock Effectively unbiased fid and inception score and where to find them.
\newblock In \emph{Proceedings of the IEEE/CVF conference on computer vision and pattern recognition}, pp.\  6070--6079, 2020.

\bibitem[Cong et~al.(2020)Cong, Zhao, Li, Wang, and Carin]{cong2020gan}
Yulai Cong, Miaoyun Zhao, Jianqiao Li, Sijia Wang, and Lawrence Carin.
\newblock Gan memory with no forgetting.
\newblock \emph{Advances in Neural Information Processing Systems}, 33:\penalty0 16481--16494, 2020.

\bibitem[Cui et~al.(2018)Cui, Song, Sun, Howard, and Belongie]{cui2018large}
Yin Cui, Yang Song, Chen Sun, Andrew Howard, and Serge Belongie.
\newblock Large scale fine-grained categorization and domain-specific transfer learning.
\newblock In \emph{Proceedings of the IEEE conference on computer vision and pattern recognition}, pp.\  4109--4118, 2018.

\bibitem[Devlin et~al.(2018)Devlin, Chang, Lee, and Toutanova]{devlin2018bert}
Jacob Devlin, Ming-Wei Chang, Kenton Lee, and Kristina Toutanova.
\newblock Bert: Pre-training of deep bidirectional transformers for language understanding.
\newblock \emph{arXiv preprint arXiv:1810.04805}, 2018.

\bibitem[Dwivedi \& Roig.(2019)Dwivedi and Roig.]{dwivedi2019}
K.~Dwivedi and G.~Roig.
\newblock Representation similarity analysis for efficient task taxonomy and transfer learning.
\newblock In \emph{{CVPR}}. {IEEE} Computer Society, 2019.

\bibitem[Elaraby et~al.(2022)Elaraby, Barakat, and Rezk]{elaraby2022conditional}
Nagwa Elaraby, Sherif Barakat, and Amira Rezk.
\newblock A conditional gan-based approach for enhancing transfer learning performance in few-shot hcr tasks.
\newblock \emph{Scientific Reports}, 12\penalty0 (1):\penalty0 16271, 2022.

\bibitem[Finn et~al.(2016)Finn, Tan, Duan, Darrell, Levine, and Abbeel]{finn2016deep}
Chelsea Finn, Xin~Yu Tan, Yan Duan, Trevor Darrell, Sergey Levine, and Pieter Abbeel.
\newblock Deep spatial autoencoders for visuomotor learning.
\newblock In \emph{Robotics and Automation (ICRA), 2016 IEEE International Conference on}, pp.\  512--519. IEEE, 2016.

\bibitem[Ge \& Yu(2017)Ge and Yu]{ge2017borrowing}
Weifeng Ge and Yizhou Yu.
\newblock Borrowing treasures from the wealthy: Deep transfer learning through selective joint fine-tuning.
\newblock In \emph{Proceedings of the IEEE conference on computer vision and pattern recognition}, pp.\  1086--1095, 2017.

\bibitem[Gulrajani et~al.(2017)Gulrajani, Ahmed, Arjovsky, Dumoulin, and Courville]{gulrajani2017improved}
Ishaan Gulrajani, Faruk Ahmed, Martin Arjovsky, Vincent Dumoulin, and Aaron~C Courville.
\newblock Improved training of wasserstein gans.
\newblock \emph{Advances in neural information processing systems}, 30, 2017.

\bibitem[Guo et~al.(2019)Guo, Shi, Kumar, Grauman, Rosing, and Feris]{guo2019spottune}
Yunhui Guo, Honghui Shi, Abhishek Kumar, Kristen Grauman, Tajana Rosing, and Rogerio Feris.
\newblock Spottune: transfer learning through adaptive fine-tuning.
\newblock In \emph{Proceedings of the IEEE/CVF conference on computer vision and pattern recognition}, pp.\  4805--4814, 2019.

\bibitem[Heusel et~al.(2017)Heusel, Ramsauer, Unterthiner, Nessler, and Hochreiter]{heusel2017gans}
Martin Heusel, Hubert Ramsauer, Thomas Unterthiner, Bernhard Nessler, and Sepp Hochreiter.
\newblock Gans trained by a two time-scale update rule converge to a local nash equilibrium.
\newblock \emph{Advances in neural information processing systems}, 30, 2017.

\bibitem[Hinton et~al.(2015)Hinton, Vinyals, and Dean]{hinton2015distilling}
Geoffrey Hinton, Oriol Vinyals, and Jeff Dean.
\newblock Distilling the knowledge in a neural network.
\newblock \emph{arXiv preprint arXiv:1503.02531}, 2015.

\bibitem[Howard \& Ruder(2018)Howard and Ruder]{howard2018universal}
Jeremy Howard and Sebastian Ruder.
\newblock Universal language model fine-tuning for text classification.
\newblock In \emph{Proceedings of the 56th Annual Meeting of the Association for Computational Linguistics (Volume 1: Long Papers)}, pp.\  328--339, 2018.

\bibitem[Kirkpatrick et~al.(2017)Kirkpatrick, Pascanu, Rabinowitz, Veness, Desjardins, Rusu, Milan, Quan, Ramalho, Grabska-Barwinska, et~al.]{kirkpatrick2017overcoming}
James Kirkpatrick, Razvan Pascanu, Neil Rabinowitz, Joel Veness, Guillaume Desjardins, Andrei~A Rusu, Kieran Milan, John Quan, Tiago Ramalho, Agnieszka Grabska-Barwinska, et~al.
\newblock Overcoming catastrophic forgetting in neural networks.
\newblock \emph{Proceedings of the national academy of sciences}, 114\penalty0 (13):\penalty0 3521--3526, 2017.

\bibitem[Krizhevsky et~al.(2009)Krizhevsky, Hinton, et~al.]{krizhevsky2009learning}
Alex Krizhevsky, Geoffrey Hinton, et~al.
\newblock Learning multiple layers of features from tiny images.
\newblock \emph{Citeseer}, 2009.

\bibitem[Le et~al.(2020)Le, Zhou, Ding, and Tarokh]{le2020supervised}
Cat~P Le, Yi~Zhou, Jie Ding, and Vahid Tarokh.
\newblock Supervised encoding for discrete representation learning.
\newblock In \emph{ICASSP 2020-2020 IEEE International Conference on Acoustics, Speech and Signal Processing (ICASSP)}, pp.\  3447--3451. IEEE, 2020.

\bibitem[Le et~al.(2021{\natexlab{a}})Le, Soltani, Ravier, and Tarokh]{le2021improved}
Cat~P Le, Mohammadreza Soltani, Robert Ravier, and Vahid Tarokh.
\newblock Improved automated machine learning from transfer learning.
\newblock \emph{arXiv e-prints}, pp.\  arXiv--2103, 2021{\natexlab{a}}.

\bibitem[Le et~al.(2021{\natexlab{b}})Le, Soltani, Ravier, and Tarokh]{le2021task}
Cat~P Le, Mohammadreza Soltani, Robert Ravier, and Vahid Tarokh.
\newblock Task-aware neural architecture search.
\newblock In \emph{ICASSP 2021-2021 IEEE International Conference on Acoustics, Speech and Signal Processing (ICASSP)}, pp.\  4090--4094. IEEE, 2021{\natexlab{b}}.

\bibitem[Le et~al.(2022{\natexlab{a}})Le, Soltani, Dong, and Tarokh]{le2022fisher}
Cat~P Le, Mohammadreza Soltani, Juncheng Dong, and Vahid Tarokh.
\newblock Fisher task distance and its application in neural architecture search.
\newblock \emph{IEEE Access}, 10:\penalty0 47235--47249, 2022{\natexlab{a}}.

\bibitem[Le et~al.(2023)Le, Dai, Johnston, Liu, Walker, and Ghanadan]{le2023improving}
Cat~P Le, Luke Dai, Michael Johnston, Yang Liu, Marilyn Walker, and Reza Ghanadan.
\newblock Improving open-domain dialogue evaluation with a causal inference model.
\newblock \emph{Diversity in Dialogue Systems: 13th International Workshop on Spoken Dialogue System Technology (IWSDS)}, 2023.

\bibitem[Le et~al.(2022{\natexlab{b}})Le, Dong, Soltani, and Tarokh]{le2022task}
Cat~Phuoc Le, Juncheng Dong, Mohammadreza Soltani, and Vahid Tarokh.
\newblock Task affinity with maximum bipartite matching in few-shot learning.
\newblock In \emph{International Conference on Learning Representations}, 2022{\natexlab{b}}.

\bibitem[LeCun et~al.(2010)LeCun, Cortes, and Burges]{lecun2010mnist}
Yann LeCun, Corinna Cortes, and CJ~Burges.
\newblock Mnist handwritten digit database.
\newblock \emph{AT\&T Labs [Online]. Available: http://yann. lecun. com/exdb/mnist}, 2:\penalty0 18, 2010.

\bibitem[Liu et~al.(2018)Liu, Wei, Lu, and Zhou]{liu2018improved}
Shaohui Liu, Yi~Wei, Jiwen Lu, and Jie Zhou.
\newblock An improved evaluation framework for generative adversarial networks.
\newblock \emph{arXiv preprint arXiv:1803.07474}, 2018.

\bibitem[Luo et~al.(2017)Luo, Zou, Hoffman, and Fei-Fei]{luolabel}
Zelun Luo, Yuliang Zou, Judy Hoffman, and Li~F Fei-Fei.
\newblock Label efficient learning of transferable representations acrosss domains and tasks.
\newblock In \emph{Advances in Neural Information Processing Systems}, pp.\  164--176, 2017.

\bibitem[Mallya \& Lazebnik(2018)Mallya and Lazebnik]{mallya2018packnet}
Arun Mallya and Svetlana Lazebnik.
\newblock Packnet: Adding multiple tasks to a single network by iterative pruning.
\newblock In \emph{Proceedings of the IEEE conference on Computer Vision and Pattern Recognition}, pp.\  7765--7773, 2018.

\bibitem[Masana et~al.(2020)Masana, Tuytelaars, and van~de Weijer]{masana2020ternary}
Marc Masana, Tinne Tuytelaars, and Joost van~de Weijer.
\newblock Ternary feature masks: continual learning without any forgetting.
\newblock \emph{arXiv preprint arXiv:2001.08714}, 4\penalty0 (5):\penalty0 6, 2020.

\bibitem[McCloskey \& Cohen(1989)McCloskey and Cohen]{mccloskey1989catastrophic}
Michael McCloskey and Neal~J Cohen.
\newblock Catastrophic interference in connectionist networks: The sequential learning problem.
\newblock In \emph{Psychology of learning and motivation}, volume~24, pp.\  109--165. Elsevier, 1989.

\bibitem[Mihalkova et~al.(2007)Mihalkova, Huynh, and Mooney]{mihalkova2007mapping}
Lilyana Mihalkova, Tuyen Huynh, and Raymond~J Mooney.
\newblock Mapping and revising markov logic networks for transfer learning.
\newblock In \emph{AAAI}, volume~7, pp.\  608--614, 2007.

\bibitem[Mirza \& Osindero(2014)Mirza and Osindero]{mirza2014conditional}
Mehdi Mirza and Simon Osindero.
\newblock Conditional generative adversarial nets.
\newblock \emph{arXiv preprint arXiv:1411.1784}, 2014.

\bibitem[Niculescu-Mizil \& Caruana(2007)Niculescu-Mizil and Caruana]{niculescu2007inductive}
Alexandru Niculescu-Mizil and Rich Caruana.
\newblock Inductive transfer for bayesian network structure learning.
\newblock In \emph{Artificial Intelligence and Statistics}, pp.\  339--346, 2007.

\bibitem[Nilsback \& Zisserman(2008)Nilsback and Zisserman]{nilsback2008automated}
Maria-Elena Nilsback and Andrew Zisserman.
\newblock Automated flower classification over a large number of classes.
\newblock In \emph{2008 Sixth Indian conference on computer vision, graphics \& image processing}, pp.\  722--729. IEEE, 2008.

\bibitem[OpenAI(2021)]{openai}
OpenAI.
\newblock Gpt-3.5.
\newblock Computer software, 2021.
\newblock URL \url{https://openai.com/blog/gpt-3-5/}.

\bibitem[Pal \& Balasubramanian(2019)Pal and Balasubramanian]{pal2019zeroshot}
Arghya Pal and Vineeth~N Balasubramanian.
\newblock Zero-shot task transfer, 2019.

\bibitem[{Pan} \& {Yang}(2010){Pan} and {Yang}]{5288526}
S.~J. {Pan} and Q.~{Yang}.
\newblock A survey on transfer learning.
\newblock \emph{IEEE Transactions on Knowledge and Data Engineering}, 22\penalty0 (10):\penalty0 1345--1359, Oct 2010.
\newblock ISSN 1041-4347.
\newblock \doi{10.1109/TKDE.2009.191}.

\bibitem[Rajasegaran et~al.(2019)Rajasegaran, Hayat, Khan, Khan, and Shao]{rajasegaran2019random}
Jathushan Rajasegaran, Munawar Hayat, Salman~H Khan, Fahad~Shahbaz Khan, and Ling Shao.
\newblock Random path selection for continual learning.
\newblock \emph{Advances in Neural Information Processing Systems}, 32, 2019.

\bibitem[Rajasegaran et~al.(2020)Rajasegaran, Khan, Hayat, Khan, and Shah]{rajasegaran2020itaml}
Jathushan Rajasegaran, Salman Khan, Munawar Hayat, Fahad~Shahbaz Khan, and Mubarak Shah.
\newblock itaml: An incremental task-agnostic meta-learning approach.
\newblock In \emph{Proceedings of the IEEE/CVF Conference on Computer Vision and Pattern Recognition}, pp.\  13588--13597, 2020.

\bibitem[Razavian et~al.(2014)Razavian, Azizpour, Sullivan, and Carlsson]{Razavian:2014:CFO:2679599.2679731}
Ali~Sharif Razavian, Hossein Azizpour, Josephine Sullivan, and Stefan Carlsson.
\newblock Cnn features off-the-shelf: An astounding baseline for recognition.
\newblock In \emph{Proceedings of the 2014 IEEE Conference on Computer Vision and Pattern Recognition Workshops}, CVPRW '14, pp.\  512--519, Washington, DC, USA, 2014. IEEE Computer Society.
\newblock ISBN 978-1-4799-4308-1.
\newblock \doi{10.1109/CVPRW.2014.131}.
\newblock URL \url{http://dx.doi.org.stanford.idm.oclc.org/10.1109/CVPRW.2014.131}.

\bibitem[Rebuffi et~al.(2017)Rebuffi, Kolesnikov, Sperl, and Lampert]{rebuffi2017icarl}
Sylvestre-Alvise Rebuffi, Alexander Kolesnikov, Georg Sperl, and Christoph~H Lampert.
\newblock icarl: Incremental classifier and representation learning.
\newblock In \emph{Proceedings of the IEEE conference on Computer Vision and Pattern Recognition}, pp.\  2001--2010, 2017.

\bibitem[Rios \& Itti(2018)Rios and Itti]{rios2018closed}
Amanda Rios and Laurent Itti.
\newblock Closed-loop memory gan for continual learning.
\newblock \emph{arXiv preprint arXiv:1811.01146}, 2018.

\bibitem[Robins(1995)]{robins1995catastrophic}
Anthony Robins.
\newblock Catastrophic forgetting, rehearsal and pseudorehearsal.
\newblock \emph{Connection Science}, 7\penalty0 (2):\penalty0 123--146, 1995.

\bibitem[Salimans et~al.(2016)Salimans, Goodfellow, Zaremba, Cheung, Radford, and Chen]{salimans2016improved}
Tim Salimans, Ian Goodfellow, Wojciech Zaremba, Vicki Cheung, Alec Radford, and Xi~Chen.
\newblock Improved techniques for training gans.
\newblock \emph{Advances in neural information processing systems}, 29, 2016.

\bibitem[Seff et~al.(2017)Seff, Beatson, Suo, and Liu]{seff2017continual}
Ari Seff, Alex Beatson, Daniel Suo, and Han Liu.
\newblock Continual learning in generative adversarial nets.
\newblock \emph{arXiv preprint arXiv:1705.08395}, 2017.

\bibitem[Silver \& Bennett(2008)Silver and Bennett]{silver2008guest}
Daniel~L Silver and Kristin~P Bennett.
\newblock Guest editor’s introduction: special issue on inductive transfer learning.
\newblock \emph{Machine Learning}, 73\penalty0 (3):\penalty0 215--220, 2008.

\bibitem[Singh et~al.(2020)Singh, Verma, Mazumder, Carin, and Rai]{singh2020calibrating}
Pravendra Singh, Vinay~Kumar Verma, Pratik Mazumder, Lawrence Carin, and Piyush Rai.
\newblock Calibrating cnns for lifelong learning.
\newblock \emph{Advances in Neural Information Processing Systems}, 33:\penalty0 15579--15590, 2020.

\bibitem[Standley et~al.(2020{\natexlab{a}})Standley, Zamir, Chen, Guibas, Malik, and Savarese]{pmlr-v119-standley20a}
Trevor Standley, Amir Zamir, Dawn Chen, Leonidas Guibas, Jitendra Malik, and Silvio Savarese.
\newblock Which tasks should be learned together in multi-task learning?
\newblock In Hal~Daumé III and Aarti Singh (eds.), \emph{Proceedings of the 37th International Conference on Machine Learning}, volume 119 of \emph{Proceedings of Machine Learning Research}, pp.\  9120--9132. PMLR, 13--18 Jul 2020{\natexlab{a}}.
\newblock URL \url{http://proceedings.mlr.press/v119/standley20a.html}.

\bibitem[Standley et~al.(2020{\natexlab{b}})Standley, Zamir, Chen, Guibas, Malik, and Savarese]{standley2020tasks}
Trevor Standley, Amir Zamir, Dawn Chen, Leonidas Guibas, Jitendra Malik, and Silvio Savarese.
\newblock Which tasks should be learned together in multi-task learning?
\newblock In \emph{International Conference on Machine Learning}, pp.\  9120--9132. PMLR, 2020{\natexlab{b}}.

\bibitem[Varshney et~al.(2021)Varshney, Verma, Srijith, Carin, and Rai]{varshney2021cam}
Sakshi Varshney, Vinay~Kumar Verma, PK~Srijith, Lawrence Carin, and Piyush Rai.
\newblock Cam-gan: Continual adaptation modules for generative adversarial networks.
\newblock \emph{Advances in Neural Information Processing Systems}, 34:\penalty0 15175--15187, 2021.

\bibitem[Vaswani et~al.(2017)Vaswani, Shazeer, Parmar, Uszkoreit, Jones, Gomez, Kaiser, and Polosukhin]{vaswani2017attention}
Ashish Vaswani, Noam Shazeer, Niki Parmar, Jakob Uszkoreit, Llion Jones, Aidan~N Gomez, {\L}ukasz Kaiser, and Illia Polosukhin.
\newblock Attention is all you need.
\newblock \emph{Advances in neural information processing systems}, 30, 2017.

\bibitem[Vaswani et~al.(2021)Vaswani, Shazeer, Parmar, Uszkoreit, Jones, Gomez, Kaiser, and Polosukhin]{vaswani2021dalle}
Ashish Vaswani, Noam Shazeer, Niki Parmar, Jakob Uszkoreit, Llion Jones, Aidan~N. Gomez, Lukasz Kaiser, and Illia Polosukhin.
\newblock {DALL}\textbullet{E}: Creating images from text.
\newblock OpenAI, 2021.
\newblock URL \url{https://openai.com/dall-e/}.

\bibitem[Verma et~al.(2021)Verma, Liang, Mehta, Rai, and Carin]{verma2021efficient}
Vinay~Kumar Verma, Kevin~J Liang, Nikhil Mehta, Piyush Rai, and Lawrence Carin.
\newblock Efficient feature transformations for discriminative and generative continual learning.
\newblock In \emph{Proceedings of the IEEE/CVF Conference on Computer Vision and Pattern Recognition}, pp.\  13865--13875, 2021.

\bibitem[Wang et~al.(2019)Wang, Wehbe, and Tarr]{wang2019neural}
Aria~Y Wang, Leila Wehbe, and Michael~J Tarr.
\newblock Neural taskonomy: Inferring the similarity of task-derived representations from brain activity.
\newblock \emph{BioRxiv}, pp.\  708016, 2019.

\bibitem[Wang et~al.(2018)Wang, Wu, Herranz, Van~de Weijer, Gonzalez-Garcia, and Raducanu]{wang2018transferring}
Yaxing Wang, Chenshen Wu, Luis Herranz, Joost Van~de Weijer, Abel Gonzalez-Garcia, and Bogdan Raducanu.
\newblock Transferring gans: generating images from limited data.
\newblock In \emph{Proceedings of the European Conference on Computer Vision (ECCV)}, pp.\  218--234, 2018.

\bibitem[Westerlund(2019)]{westerlund2019emergence}
Mika Westerlund.
\newblock The emergence of deepfake technology: A review.
\newblock \emph{Technology innovation management review}, 9\penalty0 (11), 2019.

\bibitem[Wu et~al.(2018)Wu, Herranz, Liu, Van De~Weijer, Raducanu, et~al.]{wu2018memory}
Chenshen Wu, Luis Herranz, Xialei Liu, Joost Van De~Weijer, Bogdan Raducanu, et~al.
\newblock Memory replay gans: Learning to generate new categories without forgetting.
\newblock \emph{Advances in Neural Information Processing Systems}, 31, 2018.

\bibitem[Xu \& Zhu(2018)Xu and Zhu]{xu2018reinforced}
Ju~Xu and Zhanxing Zhu.
\newblock Reinforced continual learning.
\newblock \emph{Advances in Neural Information Processing Systems}, 31, 2018.

\bibitem[Yonekura et~al.(2021)Yonekura, Miyamoto, and Suzuki]{yonekura2021inverse}
Kazuo Yonekura, Nozomu Miyamoto, and Katsuyuki Suzuki.
\newblock Inverse airfoil design method for generating varieties of smooth airfoils using conditional wgan-gp.
\newblock \emph{arXiv preprint arXiv:2110.00212}, 2021.

\bibitem[Yoon et~al.(2017)Yoon, Yang, Lee, and Hwang]{yoon2017lifelong}
Jaehong Yoon, Eunho Yang, Jeongtae Lee, and Sung~Ju Hwang.
\newblock Lifelong learning with dynamically expandable networks.
\newblock \emph{arXiv preprint arXiv:1708.01547}, 2017.

\bibitem[Zamir et~al.(2018)Zamir, Sax, Shen, Guibas, Malik, and Savarese]{zamir2018taskonomy}
Amir~R Zamir, Alexander Sax, William~B Shen, Leonidas Guibas, Jitendra Malik, and Silvio Savarese.
\newblock Taskonomy: Disentangling task transfer learning.
\newblock In \emph{2018 IEEE Conference on Computer Vision and Pattern Recognition (CVPR)}. IEEE, 2018.

\bibitem[Zenke et~al.(2017)Zenke, Poole, and Ganguli]{zenke2017continual}
Friedemann Zenke, Ben Poole, and Surya Ganguli.
\newblock Continual learning through synaptic intelligence.
\newblock In \emph{International conference on machine learning}, pp.\  3987--3995. PMLR, 2017.

\bibitem[Zhai et~al.(2019)Zhai, Chen, Tung, He, Nawhal, and Mori]{zhai2019lifelong}
Mengyao Zhai, Lei Chen, Frederick Tung, Jiawei He, Megha Nawhal, and Greg Mori.
\newblock Lifelong gan: Continual learning for conditional image generation.
\newblock In \emph{Proceedings of the IEEE/CVF international conference on computer vision}, pp.\  2759--2768, 2019.

\bibitem[Zhai et~al.(2020)Zhai, Chen, He, Nawhal, Tung, and Mori]{zhai2020piggyback}
Mengyao Zhai, Lei Chen, Jiawei He, Megha Nawhal, Frederick Tung, and Greg Mori.
\newblock Piggyback gan: Efficient lifelong learning for image conditioned generation.
\newblock In \emph{Computer Vision--ECCV 2020: 16th European Conference, Glasgow, UK, August 23--28, 2020, Proceedings, Part XXI 16}, pp.\  397--413. Springer, 2020.

\bibitem[Zhu et~al.(2017)Zhu, Zhang, Pathak, Darrell, Efros, Wang, and Shechtman]{zhu2017toward}
Jun-Yan Zhu, Richard Zhang, Deepak Pathak, Trevor Darrell, Alexei~A Efros, Oliver Wang, and Eli Shechtman.
\newblock Toward multimodal image-to-image translation.
\newblock \emph{Advances in neural information processing systems}, 30, 2017.

\end{thebibliography}
\bibliographystyle{iclr2024_conference}

\appendix
\section{Experimental Setup}\label{setup}
In this work, we construct $40$ generative tasks based on popular datasets such as MNIST~\citep{lecun2010mnist}, CIFAR-10~\citep{krizhevsky2009learning}, CIFAR-100~\citep{krizhevsky2009learning}, and Oxford Flower~\citep{nilsback2008automated}. For MNIST, we define $10$ distinct generative tasks, each focused on generating a specific digit (i.e., $0, 1, \ldots, 9$). Task $0$, for example, is designed to generate the digit $0$, while task $1$ generates the digit $1$, and so on. For the CIFAR-10 dataset, we also construct $10$ generative tasks, with each task aimed at generating a specific object category such as airplane, automobile, bird, cat, deer, dog, frog, horse, ship, and truck. Similarly, for the CIFAR-100 dataset, we create $10$ target tasks, each corresponding to a specific image class, including bear, leopard, lion, tiger, wolf, bus, pickup truck, train, streetcar, and tractor. In Oxford Flower dataset, we consider $10$ classes of flowers, including phlox, rose, calendula, iris, shasta daisy, bellflower, viola, goldquelle, peony, aquilegia. Each flower category consists of $80$ image samples. The sample was originally $128\times128$, but resized to $16\times16$ to reduce the computational complexity.

To represent the generative tasks, we utilize the conditional Wasserstein GAN with Gradient Penalty (cWGAN-GP) model~\citep{gulrajani2017improved, yonekura2021inverse}. In each experiment, we select a specific task as the target task, while considering the other tasks as source tasks. To represent these source tasks, we train the cWGAN-GP model on their respective datasets. This enables us to generate high-quality samples that are representative of the source tasks. Once trained, we can use the cWGAN-GP model as the representation network for the generative tasks. This model is then applied to our proposed mode-aware continual learning framework. We compare our method against several approaches, including individual learning~\citep{mirza2014conditional}, sequential fine-tuning~\citep{wang2018transferring}, multi-task learning~\citep{standley2020tasks}, EWC-GAN~\citep{seff2017continual}, Lifelong-GAN~\citep{zhai2019lifelong}, and CAM-GAN~\citep{varshney2021cam}. Individual learning~\citep{mirza2014conditional} involves training the cGAN model on a specific task in isolation. In sequential fine-tuning~\citep{wang2018transferring}, the cGAN model is trained sequentially on source and target tasks. Multi-task learning~\citep{standley2020tasks}, on the other hand, involves training a cGAN model on a joint dataset created from both the source and target tasks. Our method is designed to improve on these approaches by enabling the continual learning of generative tasks while mitigating catastrophic forgetting.

\section{Theoretical Analysis}\label{appendix-proof}
We first recall the definition of the GAN's discriminator loss as follows:

\begin{definition}[Discriminator Loss]
\label{def:discriminator}
Let $x=\{x_1,\ldots,x_m\}$ be the real data samples, $z$ denote the random vector, and $\theta_\mathcal{D}$ be the discriminator's parameters. $\mathcal{D}$ is trained to maximize the probability of assigning the correct label to both training real samples and generated samples $\mathcal{G}(z)$ from the generator $\mathcal{G}$. The objective of the discriminator is to maximize the following function: 
\begin{equation}
     \nabla_{\theta_\mathcal{D}} \sum_{i=1}^m \left[ \log \mathcal{D} \left( x^{(i)} \right) + \log \left( 1-\mathcal{D} \left( \mathcal{G}\left( z^{(i)} \right) \right) \right) \right]
\end{equation}
\end{definition}

We recall the definition of Fisher Information matrix (FIM)~\citep{le2022fisher} as follows:

\begin{definition}[Fisher Information]
Given dataset $X$, let $N$ denote a neural network with weights $\theta$, and the negative log-likelihood loss function $L(\theta):= L(\theta,X)$. FIM is described as follows:
\begin{align}\label{Fihermatrix}
    F(\theta) =\mathbb{E}\Big[\nabla_{\theta} L(\theta)\nabla_{\theta} L(\theta)^T\Big] = -\mathbb{E}\Big[H\big(L(\theta)\big)\Big]
\end{align}
\end{definition}

Next, we present the proof of Theorem~\ref{theorem1}.

\begin{theorem1}
Let $X_a$ be the source data, characterized by the density function $p_a$. Let $X_b$ be the data for the target mode with data density function $p_b$, $p_b \neq p_a$. 
Let $\theta$ denote the model's parameters. 
Consider the loss functions $L_a(\theta) = \mathbb{E}[l(X_a;\theta)]$ and $L_b(\theta) = \mathbb{E}[l(X_b;\theta)]$. 
Assume that both $L_a(\theta)$ and $L_b(\theta)$ are strictly convex and possess distinct global minima. 
Let $X_n$ denote the mixture data of $X_a$ and $X_b$ described by $p_n = \alpha p_a + (1-\alpha) p_b$, where $\alpha \in (0,1)$. The corresponding loss function is given by $L_n(\theta) = \mathbb{E}[l(X_n;\theta)]$.
Under these assumptions, it follows that $\theta^* = \arg \min_{\theta}L_n(\theta)$ satisfies:
\begin{equation}
    L_a(\theta^*) > \min_{\theta} L_a(\theta)
\end{equation}
\end{theorem1}

\begin{proof}[\textbf{Proof of Theorem~\ref{theorem1}}]

Assume toward contradiction that $L_a(\theta^*) > \min_{\theta} L_a(\theta)$ does not hold. Because $L_a(\theta^*) \geq \min_{\theta} L_a(\theta)$ always holds, we must have that: 
\begin{equation}\label{eqn:1st_fact}
    L_a(\theta^*) = \min_{\theta} L_a(\theta).
\end{equation}
By the linearity of expectation, we have that:
$$L_n(\theta) = \alpha L_a(\theta) + (1-\alpha) L_b(\theta)$$ 

Hence, we have
\begin{align*}
 \min_{\theta} L_n(\theta) &= L_n(\theta^*) \\
 & = \alpha L_a(\theta^*) + (1-\alpha) L_b(\theta^*) \\
 & = \alpha \min_{\theta} L_a(\theta) + (1-\alpha)L_b(\theta^*) \\
 & > \alpha \min_{\theta} L_a(\theta) + (1-\alpha)L_a(\theta^*) \\
 & = \alpha \min_{\theta} L_a(\theta) + (1-\alpha)\min_{\theta}L_a(\theta)\\
 & = \min_{\theta}L_a(\theta) 
\end{align*}
where in the third equality we use the facts that both $L_a$ and $L_b$ are strongly convex and have different global minimum. Because $L_a$ and $L_b$ have the same optimal value (assumed to be 0) and that $\theta^*$ is not the optimal point for $L_b$, we must have $L_b(\theta^*) > L_b(\theta^*_b) = L_a(\theta^*)$ where $\theta^*_b = \argmin_{\theta}L_b(\theta)$. 

Therefore, we have proved that $\min_{\theta} L_n(\theta) = L_n(\theta^*) > \min_{\theta}L_a(\theta)$, contradicting to Eq.~\eqref{eqn:1st_fact}. 
\end{proof}

\section{Ablation Studies}

\subsection{Mode-Aware Transfer Learning}
\begin{figure}
\centering
\includegraphics[width=0.95\textwidth]{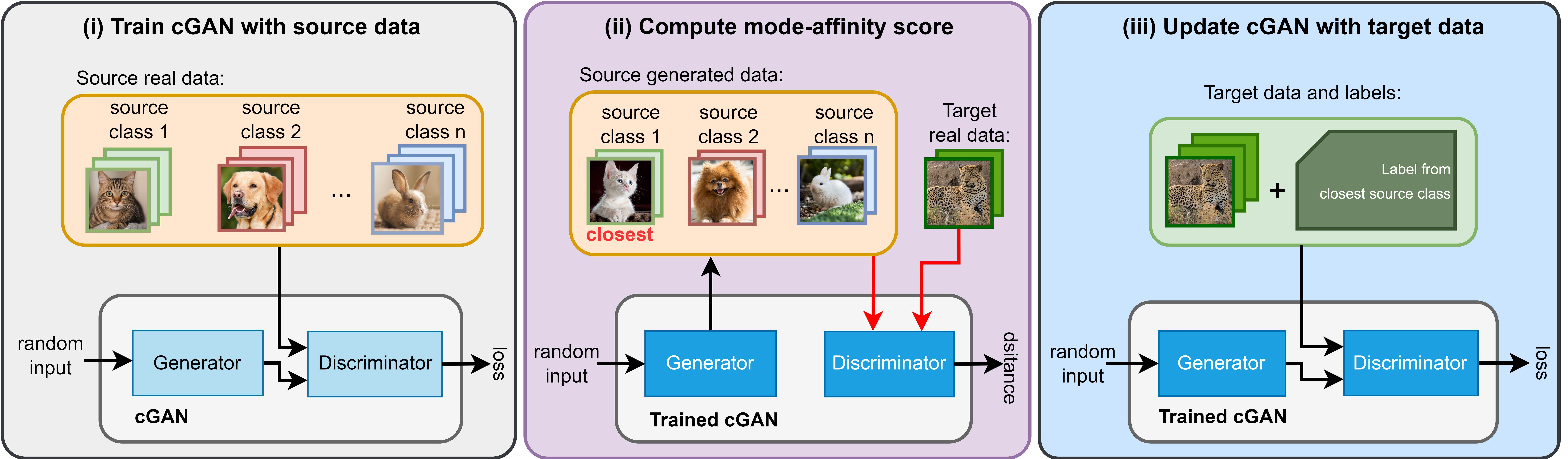}
\caption{The overview of mode-aware transfer learning framework for the conditional Generative Adversarial Network: (i) Representing source data classes using cGAN, (ii) Computing the mode-affinity from each source mode to the target, (iii) Fine-tuning the generative model using the target data and the label of the closest mode for transfer learning.}
\label{fig:transfer-learning}
\end{figure}

We apply the proposed mode-affinity score to transfer learning in an image generation scenario. The proposed similarity measure enables the identification of the closest modes or data classes to support the learning of the target mode. Here, we introduce a \textit{mode-aware transfer learning} framework that quickly adapts a pre-trained cGAN model to learn the target mode. The overview of the transfer learning framework is illustrated in Figure~\ref{fig:transfer-learning}. Particularly, we select the closest source mode from the pool of multiple learned modes based on the computed dMAS. 

To leverage the knowledge of the closest mode for training the target mode, we assign the target data samples with labels of the closest mode. Subsequently, we use these modified target data samples to fine-tune the generator and discriminator of the pre-trained cGAN model. Figure~\ref{fig:transfer-learning}(3) illustrates the transfer learning method, where the data class 1 (i.e., cat images) is the most similar to the target data (i.e., leopard image) based on the computed dMAS. Hence, we assign the label of class 1 to the leopard images. The pre-trained GAN model uses this modified target data to quickly adapt the cat image generation to the leopard image generation. The mode-aware algorithm for transfer learning in cGAN is described in Algorithm~\ref{alg2}. By assigning the closest mode's label to the target data samples, our method can effectively fine-tune the relevant parts of cGAN for learning the target mode. This approach helps improve the training process and reduces the number of required target training data.

\begin{algorithm}[t]
\SetKwInput{KwInput}{Input}         
\SetKwInput{KwOutput}{Output}       
\SetKwInput{KwData}{Data}
\SetKwFunction{FewShot}{Main}
\SetKwFunction{DTAS}{dMAS}
\SetKwFunction{MaxMatching}{MaximumMatching}
\SetKwComment{Comment}{$\triangleright$ }{}
\SetCommentSty{scriptsize}

\DontPrintSemicolon

\KwData{Source data: $(X_{train}, y_{train})$, Target data: $X_{target}$}
\KwInput{The generator $\mathcal{G}$ and discriminator $\mathcal{D}$ of cGAN}
\KwOutput{Target generator $\mathcal{G}_{\Bar{\theta}}$}

    \SetKwProg{Fn}{Function}{:}{}
    \Fn{\DTAS{$X_a, y_a, X_b, \mathcal{G}, \mathcal{D}$}}{
        Generate data $\tilde{X}_a$ of class label $y_a$ using the generator $\mathcal{G}$\;
        Compute $H_{a}$ from the loss of discriminator $\mathcal{D}$ using $\{X_a, \tilde{X}_a\}$\;
        Compute $H_{b}$ from the loss of discriminator $\mathcal{D}$ using $\{X_b, \tilde{X}_a\}$\;
        \KwRet $\displaystyle s[a, b] = \frac{1}{\sqrt{2}} \norm{H_{a}^{1/2} - H_{b}^{1/2}}_F$
    }
    
    \SetKwProg{Fn}{Function}{:}{} \Fn{\FewShot}{ 
        Train ($\mathcal{G}_{\theta}$, $\mathcal{D}_{\theta}$) with $X_{train}, y_{train}$  \Comment*[r]{Pre-train cGAN model}
        Construct S source modes, each from a data class in $y_{train}$\;

        \For{$i=1,2,\ldots,S$}{ 
            $s_i = \DTAS(X_{train_i}, y_{train_i}, X_{target}, \mathcal{G}_{\theta}, \mathcal{D}_{\theta})$ \Comment*[r]{Find the closest modes}
        }
        \KwRet closest mode(s): $i^* = \underset{i}{\mathrm{argmin}}\ s_i$\;

        \Comment*[r]{Fine-tune with the target task}
        \While{$\theta$ not converged}{
            Update $\mathcal{G}_{\theta}$, $\mathcal{D}_{\theta}$ using real data $X_{target}$ and closest source label $y_{train_{i^*}}$\;
        }
        
        \KwRet $\mathcal{G}_{\Bar{\theta}}$\;

    }

\caption{Mode-Aware Transfer Learning for Conditional Generative Adversarial Networks}
\label{alg2}
\end{algorithm}


Next, we conduct experiments employing mode affinity scores within the context of transfer learning scenarios. These experiments were designed to assess the effectiveness of our proposed mode-affinity measure in the transfer learning framework. In this scenario, each generative task corresponds to a single data class within the MNIST~\citep{lecun2010mnist}, CIFAR-10~\citep{krizhevsky2009learning}, and CIFAR-100~\citep{krizhevsky2009learning} datasets. Here, in our transfer learning framework, we leverage the computed mode-affinity scores between generative tasks. Specifically, we utilize this distance metric to identify the mode closest to the target mode and then fine-tune the conditional Generative Adversarial Network (cGAN) accordingly. To achieve this, we assign the target data samples with the labels of the closest mode and use these newly-labeled samples to train the cGAN model. By doing so, the generative model can benefit from the knowledge acquired from the closest mode, enabling quick adaptation in learning the target mode. In this study, we compare our proposed transfer learning framework with several baselines and state-of-the-art approaches, including individual learning~\citep{mirza2014conditional}, sequential fine-tuning~\citep{wang2018transferring}, multi-task learning~\citep{standley2020tasks}, and FID-transfer learning~\citep{wang2018transferring}. Additionally, we present a performance comparison of our mode-aware transfer learning approach with these methods for 10-shot, 20-shot, and 100-shot scenarios in the MNIST, CIFAR-10, and CIFAR-100 datasets (i.e., the target dataset contains only 10, 20, or 100 data samples).

\begin{table}[t]
\caption{Comparison of the mode-aware transfer learning framework for cGAN against other baselines and FID-transfer learning approach in terms of FID.}
\label{transfer-learning-table}
\begin{center}
\begin{tabular}{l|c|ccc}
\hline
\multicolumn{1}{l}{} &\multicolumn{1}{c}{} &\multicolumn{3}{c}{\textbf{MNIST}} \\
\multicolumn{1}{l}{\textbf{Approach}} &\multicolumn{1}{c}{\textbf{Target}} &\multicolumn{1}{c}{\textbf{10-shot}} &\multicolumn{1}{c}{\textbf{20-shot}} &\multicolumn{1}{c}{\textbf{100-shot}} \\
\hline
Individual Learning~\citep{mirza2014conditional}           
& Digit 0  & 34.25 & 27.17 & 19.62 \\
Sequential Fine-tuning~\citep{zhai2019lifelong}           
& Digit 0  & 29.68 & 24.22 & 16.14 \\
Multi-task Learning~\citep{standley2020tasks}             
& Digit 0  & 26.51 & 20.74 & 10.95 \\
FID-Transfer Learning~\citep{wang2018transferring}             
& Digit 0  & \textbf{12.64} & \textbf{7.51}  & \textbf{5.53} \\
\hline
\textbf{MA-Transfer Learning (ours)}      
& \textbf{Digit 0}  & \textbf{12.64} & \textbf{7.51}  & \textbf{5.53} \\

\hline
Individual Learning~\citep{mirza2014conditional}           
& Digit 1  & 35.07 & 29.62  & 20.83 \\
Sequential Fine-tuning~\citep{zhai2019lifelong}           
& Digit 1  & 28.35 & 24.79 & 15.85 \\
Multi-task Learning~\citep{standley2020tasks}             
& Digit 1  & 26.98 & 21.56  & 10.68 \\
FID-Transfer Learning~\citep{wang2018transferring}             
& Digit 1  & \textbf{11.35} & \textbf{7.12}  & \textbf{5.28} \\
\hline
\textbf{MA-Transfer Learning (ours)}      
& \textbf{Digit 1}  & \textbf{11.35} & \textbf{7.12}  & \textbf{5.28} \\

\hline

\multicolumn{1}{l}{} &\multicolumn{1}{c}{} &\multicolumn{3}{c}{\textbf{CIFAR-10}} \\
\multicolumn{1}{l}{\textbf{Approach}} &\multicolumn{1}{c}{\textbf{Target}} &\multicolumn{1}{c}{\textbf{10-shot}} &\multicolumn{1}{c}{\textbf{20-shot}} &\multicolumn{1}{c}{\textbf{100-shot}} \\
\hline
Individual Learning~\citep{mirza2014conditional}           
& Truck  & 89.35 & 81.74  & 72.18 \\
Sequential Fine-tuning~\citep{zhai2019lifelong}           
& Truck  & 76.93 & 70.39  & 61.41 \\
Multi-task Learning~\citep{standley2020tasks}             
& Truck  & 72.06 & 65.38  & 55.29 \\
FID-Transfer Learning~\citep{wang2018transferring}             
& Truck  & \textbf{51.05} & \textbf{44.93}  & \textbf{36.74} \\
\hline
\textbf{MA-Transfer Learning (ours)}      
& \textbf{Truck}  & \textbf{51.05} & \textbf{44.93}  & \textbf{36.74} \\

\hline
Individual Learning~\citep{mirza2014conditional}                 
& Cat  & 80.25 & 74.46  & 65.18 \\
Sequential Fine-tuning~\citep{zhai2019lifelong}           
& Cat  & 73.51 & 68.23 & 59.08 \\
Multi-task Learning~\citep{standley2020tasks}             
& Cat  & 68.73 & 61.32  & 50.65 \\
FID-Transfer Learning~\citep{wang2018transferring}             
& Cat  & \textbf{47.39} & \textbf{40.75}  & \textbf{32.46} \\
\hline
\textbf{MA-Transfer Learning (ours)}      
& \textbf{Cat}  & \textbf{47.39} & \textbf{40.75}  & \textbf{32.46} \\

\hline

\multicolumn{1}{l}{} &\multicolumn{1}{c}{} &\multicolumn{3}{c}{\textbf{CIFAR-100}} \\
\multicolumn{1}{l}{\textbf{Approach}} &\multicolumn{1}{c}{\textbf{Target}} &\multicolumn{1}{c}{\textbf{10-shot}} &\multicolumn{1}{c}{\textbf{20-shot}} &\multicolumn{1}{c}{\textbf{100-shot}} \\
\hline
Individual Learning~\citep{mirza2014conditional}           
& Lion  & 87.91 & 80.21  & 72.58 \\
Sequential Fine-tuning~\citep{zhai2019lifelong}           
& Lion  & 77.56 & 70.76  & 61.33 \\
Multi-task Learning~\citep{standley2020tasks}             
& Lion  & 71.25 & 67.84  & 56.12 \\
FID-Transfer Learning~\citep{wang2018transferring}             
& Lion  & \textbf{51.08} & \textbf{46.97}  & \textbf{37.51} \\
\hline
\textbf{MA-Transfer Learning (ours)}      
& \textbf{Lion}  & \textbf{51.08} & \textbf{46.97}  & \textbf{37.51} \\

\hline
Individual Learning~\citep{mirza2014conditional}           
& Bus  & 94.82 & 89.01  & 78.47 \\
Sequential Fine-tuning~\citep{zhai2019lifelong}           
& Bus  & 88.03 & 79.51 & 67.33 \\
Multi-task Learning~\citep{standley2020tasks}             
& Bus  & 80.06 & 76.33  & 61.59 \\
FID-Transfer Learning~\citep{wang2018transferring}             
& Bus  & 61.34 & 54.18  & 46.37 \\
\hline
\textbf{MA-Transfer Learning (ours)}      
& \textbf{Bus}  & \textbf{57.16} & \textbf{50.06}  & \textbf{41.81} \\

\hline

\end{tabular}
\end{center}
\end{table}

Across all three datasets, our results demonstrate the effectiveness of our approach in terms of generative performance and its ability to efficiently learn new tasks. Our proposed framework significantly outperforms individual learning and sequential fine-tuning while demonstrating strong performance even with fewer samples compared to multi-task learning. Moreover, our approach is competitive with FID transfer learning, where the similarity measure between generative tasks is based on FID scores. Notably, our experiments with the CIFAR-100 dataset reveal that FID scores may not align with intuition and often result in poor performance. Notably, for the MNIST dataset, we consider generating digits $0$ and $1$ as the target modes. As shown in Table~\ref{transfer-learning-table}, our method outperforms individual learning, sequential fine-tuning, and multi-task learning approaches significantly, while achieving similar results compared with the FID transfer learning method. Since the individual learning model lacks training data, it can only produce low-quality samples. On the other hand, the sequential fine-tuning and multi-task learning models use the entire source dataset while training the target mode, which results in better performance than the individual learning method. However, they cannot identify the most relevant source mode and data, thus, making them inefficient compared with our proposed mode-aware transfer learning approach. In other words, the proposed approach can generate high-quality images with fewer target training samples. Notably, the proposed approach can achieve better results using only 20\% of data samples. For more complex tasks, such as generating cat and truck images in CIFAR-10 and lion and bus images in CIFAR-100, our approach achieves competitive results to other methods while requiring only 10\% training samples. Hence, the mode-aware transfer learning framework using the Discriminator-based Mode Affinity Score can effectively identify relevant source modes and utilize their knowledge for learning the target mode.

\subsection{Choice of closest modes}
In this experiment, we evaluate the effectiveness of our proposed continual learning framework by varying the number of closest existing modes used for fine-tuning the target mode. Throughout this paper, we opt to utilize the top-2 closest modes, a choice driven by its minimal computational requirements. Opting for a single closest mode (i.e., transfer learning scenarios) would essentially replace that mode with the target mode, negating the concept of continual learning. Here, we explore different scenarios across the MNIST, CIFAR-10, and CIFAR-100 datasets, where we investigate the top-2, top-3, and top-4 closest modes for continual learning. As detailed in Table~\ref{table:closest-mode}, selecting the three closest modes yields the most favorable target generation performance in the MNIST and CIFAR-10 experiments. Notably, knowledge transfer from the four closest modes results in the weakest performance. This discrepancy can be attributed to the simplicity of these datasets and their highly distinguishable data classes. In such cases, employing more tasks resembles working with dissimilar tasks, leading to negative transfer during target mode training. Conversely, in the CIFAR-100 experiment, opting for the top-4 modes yields the best performance. This outcome stems from the dataset's complexity, where utilizing a larger set of relevant modes confers an advantage during the fine-tuning process. 

In summary, the choice of the top-N closest modes is highly dependent on the dataset and available computational resources. Employing more modes necessitates significantly more computational resources and training time for memory replay of existing tasks. It's crucial to note that with an increased number of related modes, the model requires more time and data to converge effectively.

\begin{table}[t]
\caption{Comparison of the mode-aware continual learning performance between different choices of the number of closest modes}
\label{table:closest-mode}
\begin{center}
\begin{tabular}{l|cc|c}
\hline
\multicolumn{1}{l}{\textbf{Approach}} &\multicolumn{1}{c}{\textbf{Dataset}} &\multicolumn{1}{c}{\textbf{Target}} &\multicolumn{1}{c}{\textbf{Performance}}\\
\hline
MA-Continual Learning with top-2 closest modes & MNIST & Digit 0  & 6.32 \\
\textbf{MA-Continual Learning with top-3 closest modes} & \textbf{MNIST} & \textbf{Digit 0}  & \textbf{6.11} \\
MA-Continual Learning with top-4 closest modes & MNIST & Digit 0  & 6.78 \\
\hline
MA-Continual Learning with top-2 closest modes & CIFAR-10 & Truck  & 35.57 \\
\textbf{MA-Continual Learning with top-3 closest modes} & \textbf{CIFAR-10} & \textbf{Truck}  & \textbf{35.52} \\
MA-Continual Learning with top-4 closest modes & CIFAR-10 & Truck  & 36.31 \\
\hline
MA-Continual Learning with top-2 closest modes & CIFAR-100 & Lion  & 38.73 \\
MA-Continual Learning with top-3 closest modes & CIFAR-100 & Lion  & 38.54 \\
\textbf{MA-Continual Learning with top-4 closest modes} & \textbf{CIFAR-100} & \textbf{Lion}  & \textbf{38.31} \\
\hline
\end{tabular}
\end{center}
\end{table}

\begin{figure}
\centering
    \begin{subfigure}{0.49\textwidth}
    \centering
    \includegraphics[width=0.83\textwidth]{Figures/mnist_generative_mean.pdf}
    \end{subfigure}
    \begin{subfigure}{0.49\textwidth}
    \centering
    \includegraphics[width=0.85\textwidth]{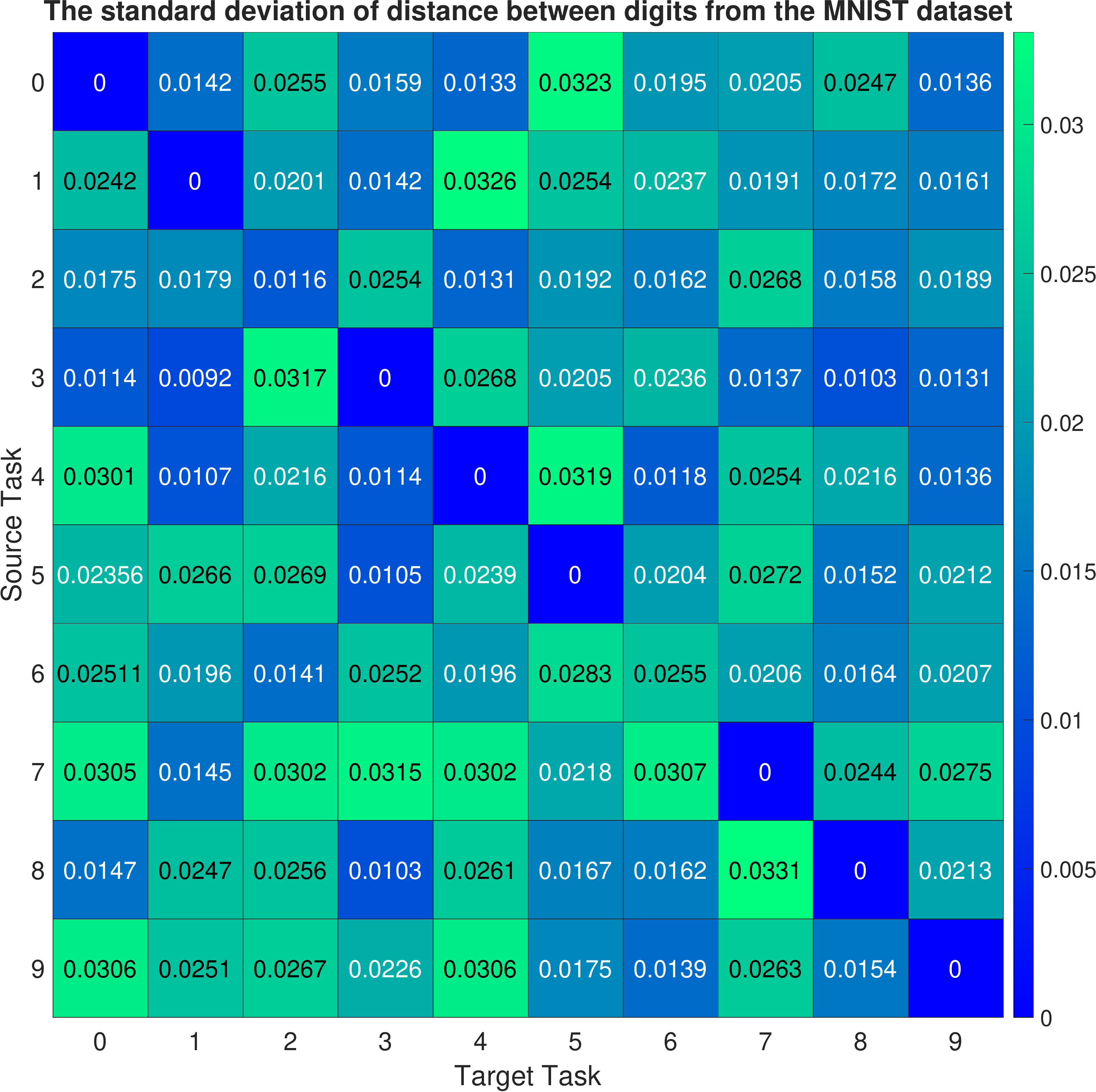}
    \end{subfigure}
\caption{The mean (left) and standard deviation (right) of computed mode-affinity scores between data classes (i.e., digits $0, 1, \ldots, 9$) of the MNIST dataset using cGAN.}
\label{fig:mnist-distance-var}
\end{figure}

\begin{figure}
\centering
    \begin{subfigure}{0.49\textwidth}
    \centering
    \includegraphics[width=0.94\textwidth]{Figures/cifar_generative_mean.pdf}
    \end{subfigure}
    \begin{subfigure}{0.49\textwidth}
    \centering
    \includegraphics[width=0.94\textwidth]{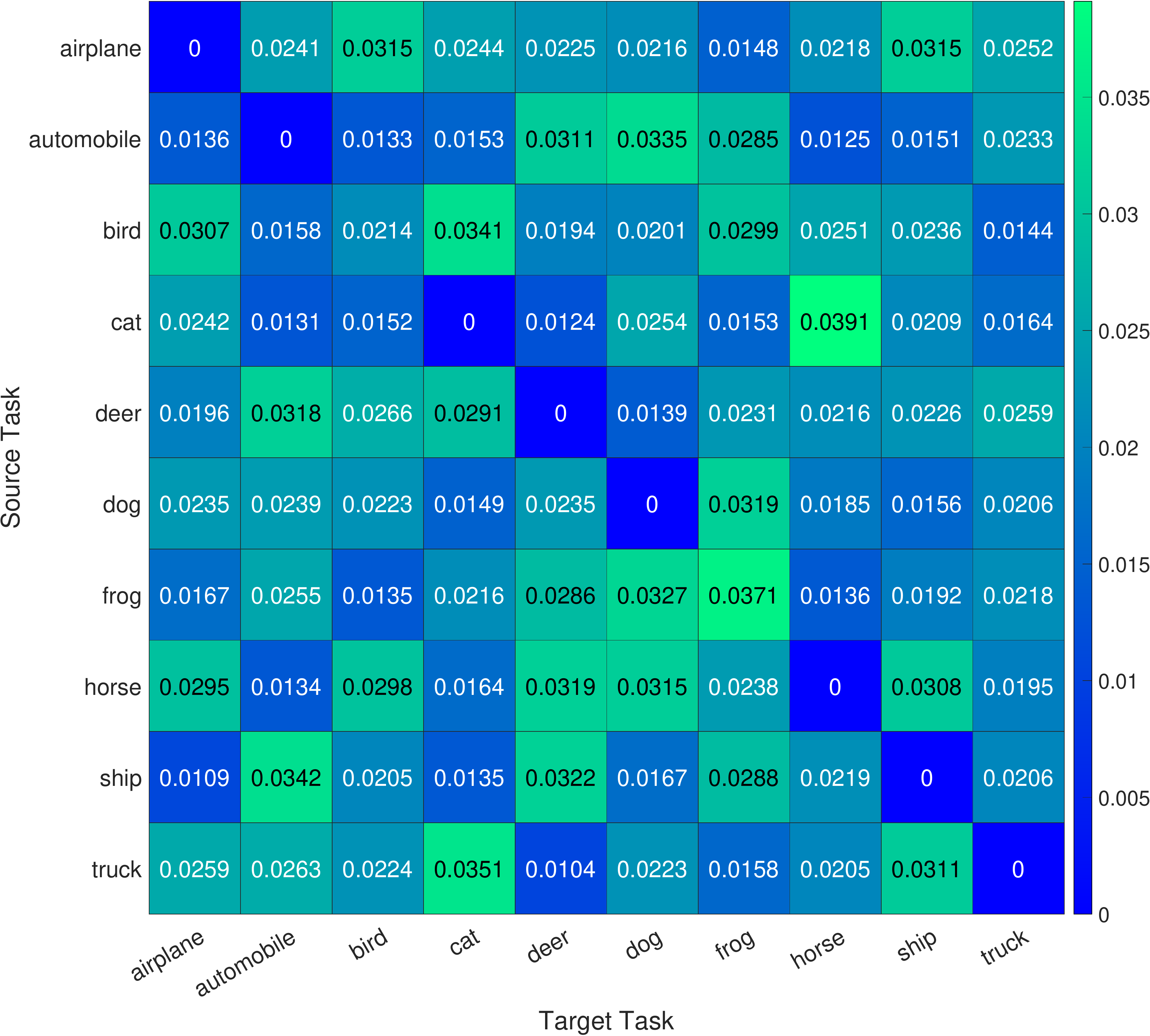}
    \end{subfigure}
\caption{The mean (left) and standard deviation (right) of computed mode-affinity scores between data classes (i.e., airplane, automobile, bird,..., truck) of the CIFAR-10 dataset using cGAN.}
\label{fig:cifar-distance-var}
\end{figure}

\begin{figure}
\centering
    \begin{subfigure}{0.49\textwidth}
    \centering
    \includegraphics[width=0.94\textwidth]{Figures/cifar100_generative_mean.pdf}
    \end{subfigure}
    \begin{subfigure}{0.49\textwidth}
    \centering
    \includegraphics[width=0.94\textwidth]{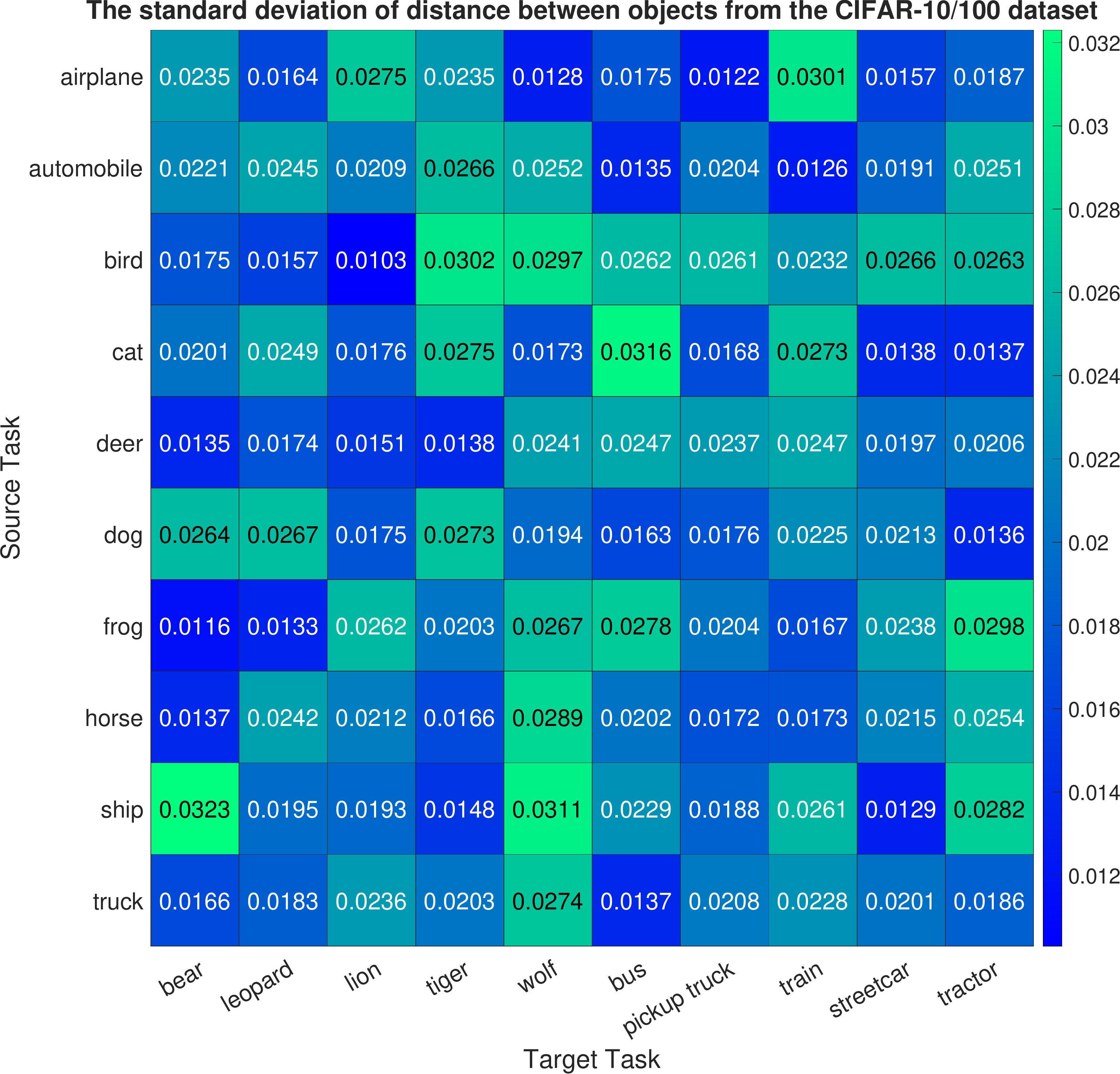}
    \end{subfigure}
\caption{The mean (left) and standard deviation (right) of computed mode-affinity scores between data classes of the CIFAR-10 and CIFAR-100 datasets using cGAN.}
\label{fig:cifar100-distance-var}
\end{figure}

\subsection{Mode Affinity Score Consistency}
To rigorously evaluate and demonstrate the statistical significance of our proposed dMAS in the context of the MNIST, CIFAR-10, and CIFAR-100 datasets, we conducted a comprehensive series of distance consistency experiments. This experimental approach was designed to validate the effectiveness and reliability of the dMAS metric in assessing the affinity between generative tasks within the realm of conditional Generative Adversarial Networks (cGANs). We first initiated each experiment by computing the dMAS distance between different tasks, ensuring a diverse range of task combinations to comprehensively test the metric's robustness. Importantly, we repeated each experiment a total of 10 times, introducing variability through distinct initialization settings for pre-training the cGAN model. This multi-run approach was adopted to account for any potential variability introduced by the initialization process and to capture the metric's performance across a spectrum of scenarios.

We calculated both the mean and standard deviation values of the dMAS distances obtained from each of these 10 experimental runs. This analysis allowed us to quantitatively assess the central tendency and variability of the dMAS measurements. To visually convey the results and facilitate a clear understanding of our findings, we illustrated the mean and standard deviation values of the dMAS distances in Figure~\ref{fig:mnist-distance-var} for the MNIST dataset, Figure~\ref{fig:cifar-distance-var} for CIFAR-10, and Figure~\ref{fig:cifar100-distance-var} for CIFAR-100. These figures provide a visual representation of the consistency and stability of the dMAS metric across different tasks and initialization settings. In summary, our experiments and comprehensive analysis conclusively establish the dMAS as a consistently reliable distance metric for assessing the affinity between generative tasks within the framework of cGANs. These findings underscore the metric's robustness and its potential utility in various generative modeling applications.

\end{document}